\documentclass[10pt,twocolumn,letterpaper]{article}
\pdfoutput=1
\usepackage{cvpr}              % To produce the CAMERA-READY version
\usepackage{graphicx}
\usepackage{amsmath}
\usepackage{amssymb}
\usepackage{booktabs}
\usepackage{color}
\usepackage[table,xcdraw]{xcolor}
\usepackage{multirow}
\usepackage{algorithm}
\usepackage{algorithmic}
\usepackage[accsupp]{axessibility}  % Improves PDF readability for those with disabilities.
\usepackage[pagebackref,breaklinks,colorlinks]{hyperref}

\usepackage[capitalize]{cleveref}
\crefname{section}{Sec.}{Secs.}
\Crefname{section}{Section}{Sections}
\Crefname{table}{Table}{Tables}
\crefname{table}{Tab.}{Tabs.}

\def\cvprPaperID{10321} % *** Enter the CVPR Paper ID here
\def\confName{CVPR}
\def\confYear{2023}

\begin{document}

%%%%%%%%% TITLE - PLEASE UPDATE
\title{Contrastive Semi-supervised  Learning for Underwater Image Restoration via Reliable Bank}

\author{Shirui
Huang\textsuperscript{1}\textsuperscript{*}, Keyan Wang\textsuperscript{1}\textsuperscript{*}\textsuperscript{$\dagger$}, Huan Liu\textsuperscript{2}, Jun Chen\textsuperscript{2},  Yunsong Li\textsuperscript{1}
\\
\textsuperscript{1} Xidian University,
\textsuperscript{2} McMaster University,
\textsuperscript{*} Equal Contribution,
\textsuperscript{$\dagger$} Corresponding Author\footnotetext{1}\\
\texttt{\small srhuang@stu.xidian.edu.cn,\{kywang, ysli\}@mail.xidian.edu.cn,\{liuh127, chenjun\}@mcmaster.ca}
}
\maketitle
\footnotetext[1]{This work is supported by the Nature Science Foundation of Shaanxi Province of China (2021JM-125).}
%%%%%%%%% ABSTRACT
\begin{abstract}
Despite the remarkable achievement of recent underwater image restoration techniques, the lack of labeled data has become a major hurdle for further progress.  
In this work, we propose a mean-teacher based \textbf{Semi}-supervised \textbf{U}nderwater \textbf{I}mage \textbf{R}estoration (\textbf{Semi-UIR}) framework to incorporate the unlabeled data into network training. However, the naive mean-teacher method suffers from two main problems: (1) The consistency loss used in training might become ineffective when the teacher's prediction is wrong. (2) Using L1 distance may cause the network to overfit wrong labels, resulting in confirmation bias.
To address the above problems, we first introduce a reliable bank to store the ``best-ever" outputs as pseudo ground truth. To assess the quality of outputs, we conduct an empirical analysis based on the monotonicity property to select the most trustworthy NR-IQA method. Besides, in view of the confirmation bias problem, we incorporate contrastive regularization to prevent the overfitting on wrong labels.
Experimental results on both full-reference and non-reference underwater benchmarks demonstrate that our algorithm has obvious improvement over SOTA methods quantitatively and qualitatively. Code has been released at https://github.com/Huang-ShiRui/Semi-UIR. 

\end{abstract}

%%%%%%%%% BODY TEXT
\section{Introduction}
\begin{figure}[!t]
\centering
\includegraphics[width=\linewidth]{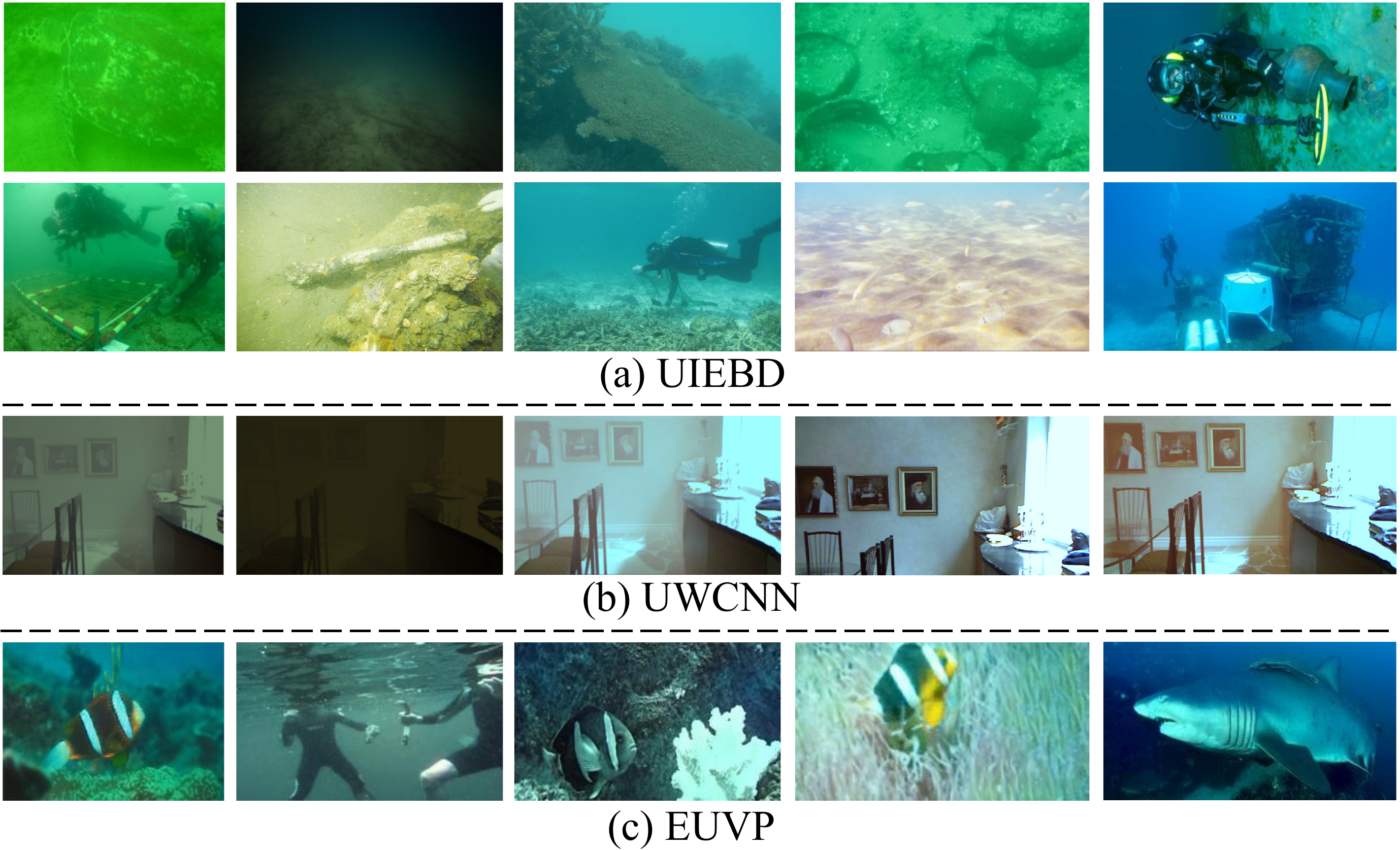}
\caption{Examples from different benchmarks. (a) shows real-world underwater images from UIEB \cite{UIEBD} with degraded images (first and second row). 
(b) shows the UWCNN training set \cite{Anwar2019UWCNN} (synthesized based on the image formation model) and (c) shows the EUVP dataset \cite{EUVP-FUnieGan} (synthesized by GAN). The ambient light and color cast of (b) and (c) are quite different from that of (a).\label{dataset1}}
\end{figure}   
Due to light refraction, absorption and scattering in underwater scenes, images taken in the water usually suffer severely from color distortion, low contrast and blur. Images with these defects tend to be less visually appealing and can potentially hinder the well-functioning of underwater robotic systems. 
% Recently, many deep learning based methods\cite{UIEBD,EUVP-FUnieGan,ucolor,cwr} have been proposed to address this problem. 
Recently, many deep learning based methods \cite{huan1,huan2,huan3,huan4,huan5} have been proposed to address image restoration problems. Numerous efforts have also been devoted to the specific domain of underwater image restoration \cite{UIEBD, EUVP-FUnieGan, ucolor, cwr, yin2021fmsnet}.
Compared with traditional methods that mostly rely on hand-crafted priors, deep learning based solutions are able to deliver superior restoration results due to their data-driven nature.

Despite their success, most of deep learning based methods are designed to learn the restoration mapping on paired datasets in a supervised manner. 
As is known, it is extremely hard, if not impossible, to acquire paired underwater images in real scenes. The existing datasets for underwater image restoration have several non-negligible issues: (1) \textit{Lack of real data.} A popular way to construct paired datasets is to synthesize underwater images using some physical model\cite{Anwar2019UWCNN} or GAN\cite{EUVP-FUnieGan}. However, there is a significant discrepancy between synthesized and real data \cite{UWNR}. As is shown in Fig. \ref{dataset1}, the ambient light and color cast of synthetic data are quite different from the real counterparts. Due to domain shift, models trained on synthetic datasets often exhibit poor generalization in real scenes.
Another way\cite{UIEBD} is to manually construct pseudo labels by selecting the best results among those produced by traditional algorithms. Inevitably, the quality of the pseudo ground truth is restricted by the restoration capability of traditional algorithms.
(2) \textit{Limited data size.} The current benchmarks only provide a very limited amount of paired data. For example, in UIEB\cite{UIEBD}, there are only 890 paired underwater images. Models learned on such a small dataset may run the risk of overfitting. In comparison, the standard datasets for image classification,  such as ImageNet\cite{imagenet}, are several orders of magnitude larger in size.

On the other hand, unlabeled underwater images are relatively easy to collect. The challenge is how to make effective use of these unlabeled data. Semi-supervised learning, which capitalizes on both labeled and unlabeled data for model training, is best suited in this kind of scenarios.
This motivates us to propose a semi-supervised scheme with the goal of improving the generalization of the resulting model on real-world underwater images. 
To be specific, we adopt the mean teacher method\cite{tarvainen2017mean} as the basis. The mean teacher method finds a way to obtain pseudo labels for unlabeled data and utilizes a consistency loss to improve the accuracy and robustness of the network. Specifically, it constructs a teacher model with improved performance from a student model via the exponential moving average (EMA) strategy. The teacher's prediction serves as the pseudo label to guide the training of the student. However, it is a non-trivial task to tailor the mean teacher method to the underwater image restoration problem. The reasons are as follows: (1) There is no guarantee that the teacher can consistently outperform the student. Wrong pseudo labels may jeopardize the training of the student network. (2) The commonly used consistency loss is based on L1 distance. The ``strict" L1 loss can easily make the model overfit wrong predictions, resulting in confirmation bias.

% To address the first issue, we construct a reliable bank to archive the best-ever outputs from the teacher as pseudo labels. The main challenge here is how to determine what are the ``best-ever" outputs? Intuitively,  non-reference image quality assessment (NR-IQA) can be leveraged to evaluate the quality of each output. However, as is noted in \cite{berman2020underwater, UIEBD}, the current NR-IQA metrics for underwater images are, to some extent, inconsistent with human visual perception.
% To identify the right one for our purpose, we compare several NR-IQA metrics using the monotonicity property as the reliability criterion. 
% Our empirical analysis suggests MUSIQ\cite{ke2021musiq} best meets the criterion. 
% For the second issue, we introduce contrastive learning as a supplementary regularization to alleviate overfitting. Unlike those conventional loss functions that are only concerned with how close the outputs and ground truths are, contrastive loss provides additional supervision to prevent the degradation of the outputs. In this sense, contrastive regularization is ideally suited to our  semi-supervised learning framework since we only have access to the degraded underwater images in the unlabeled dataset. It enables the model to take advantage of unlabeled data. 
To address the first issue, we construct a reliable bank to archive the best-ever outputs from the teacher as pseudo labels. The main challenge here is how to determine what are the ``best-ever" outputs? Intuitively,  non-reference image quality assessment (NR-IQA) can be leveraged to evaluate the quality of each output. However, as noted in \cite{berman2020underwater, UIEBD, guo2023uranker}, the current NR-IQA metrics for underwater images are, to some extent, inconsistent with human visual perception.
To identify the right one for our purpose, we compare several NR-IQA metrics using the monotonicity property as the reliability criterion. 
Our empirical analysis suggests MUSIQ\cite{ke2021musiq} best meets the criterion. 
For the second issue, we introduce contrastive learning as a supplementary regularization to alleviate overfitting. Unlike those conventional loss functions that are only concerned with how close the outputs and ground truths are, contrastive loss provides additional supervision to prevent the degradation of the outputs. In this sense, contrastive regularization is ideally suited to our  semi-supervised learning framework since we only have access to the degraded images in the unlabeled dataset. It enables the model to take advantage of unlabeled data. 

In summary, our main contributions are as follows:
(1) We propose a mean teacher based semi-supervised underwater image restoration framework named Semi-UIR, which effectively leverages the knowledge from unlabeled data to improve the generalization of the trained model on real-world data.
(2) We evaluate teacher outputs by a judiciously chosen NR-IQA metric and build a reliable bank to store best-ever teacher outputs, which ensures the reliability of pseudo-labels. 
(3) We adopt contrastive loss as a form of regularization 
to alleviate confirmation bias.
(4) Extensive experimental results demonstrate the effectiveness of our proposed methods.

\section{Related Work}
\subsection{Underwater Image Restoration Methods}
Traditional underwater image restoration methods can be categorized into model-based and model-agnostic methods.  Model-based methods\cite{galdran2015rcp, peng2018gdcp, berman2020underwater} use hand-crafted priors to estimate unknown parameters of underwater imaging model\cite{schechner2004clear}, such as transmission and ambient light.
In contrast, model-agnostic methods rely on the design of appropriate image enhancement techniques such as CLAHE\cite{clahe-mix}, Retinex\cite{labmsr}, fusion\cite{fusion-18} and MMLE\cite{MMLE}. Despite their success, the traditional methods usually fail to cope with complex real scenes. 

Most of early deep learning based underwater image restoration methods\cite{wang2019underwater,kar2021zero} accomplish the goal by exploiting physical imaging models. Specifically, they make use of neural networks to estimate the transmission and ambient light. However, inaccurate estimation of these parameters hinders such methods from achieving good performance. Recent years have seen many deep learning based methods that directly learn the restoration mapping from the labeled dataset in a supervised manner without resorting to imaging models. 
\cite{UIEBD} employs an effective network to fuse three feature maps enhanced by traditional model-agnostic methods. \cite{ucolor} designs a multi-color space encoder and a transmission-guided decoder by leveraging the ideas from  traditional model-based approaches. \cite{prwnet} proposes a wavelet boost learning strategy, through which features in the frequency domain are utilized for fine detail restoration. \cite{EUVP-FUnieGan} introduces an end-to-end network based on GAN for the purpose of real-time inference.

\subsection{Semi-supervised Learning}

In recent years, semi-supervised learning \cite{zhu2005semi} has played an increasingly important role in tackling  computer vision problems. It focuses on making effective use of both labeled and unlabeled data. Many semi-supervised methods have been developed, such as mean teacher\cite{tarvainen2017mean}, virtual adversarial learning\cite{miyato2018virtual} and FixMatch\cite{sohn2020fixmatch}. 
% Among them, the mean teacher method\cite{tarvainen2017mean}, which is based on consistency regularization, has achieved remarkable success in semi-supervised image recognition. This success also triggers its applications to other vision tasks such as semantic segmentation \cite{nips2021ael,cvpr2022psmt,cvpr2022u2pl} and image restoration \cite{acmm2021dmt, 2022semisr}. Unfortunately, to the best of our knowledge, semi-supervised learning is rarely explored in underwater image restoration.
% \cite{20-semi} makes an initial attempt in this direction
% by training a single network with both supervised and unsupervised losses.  In comparison, we adopt a more systematic approach and introduce several effective techniques in handling unlabeled data, including mean teacher, reliable bank and contrastive loss.
Among them, the mean teacher method\cite{tarvainen2017mean}, which is based on consistency regularization, has achieved
remarkable success in semi-supervised image recognition. This success also triggers its applications to
other vision tasks such as semantic segmentation \cite{nips2021ael,cvpr2022psmt,cvpr2022u2pl} and image restoration \cite{acmm2021dmt, 2022semisr}. Unfortunately, to the best of our knowledge, semi-supervised learning is rarely explored in underwater image restoration.
\cite{20-semi} makes an initial attempt in this direction
by training a single network with both supervised and unsupervised losses.  In comparison, we adopt a more systematic approach and introduce several techniques in handling unlabeled data, including mean teacher, reliable bank and contrastive loss.

\begin{figure*}[!t]
\centering
\includegraphics[width=0.8 \linewidth]{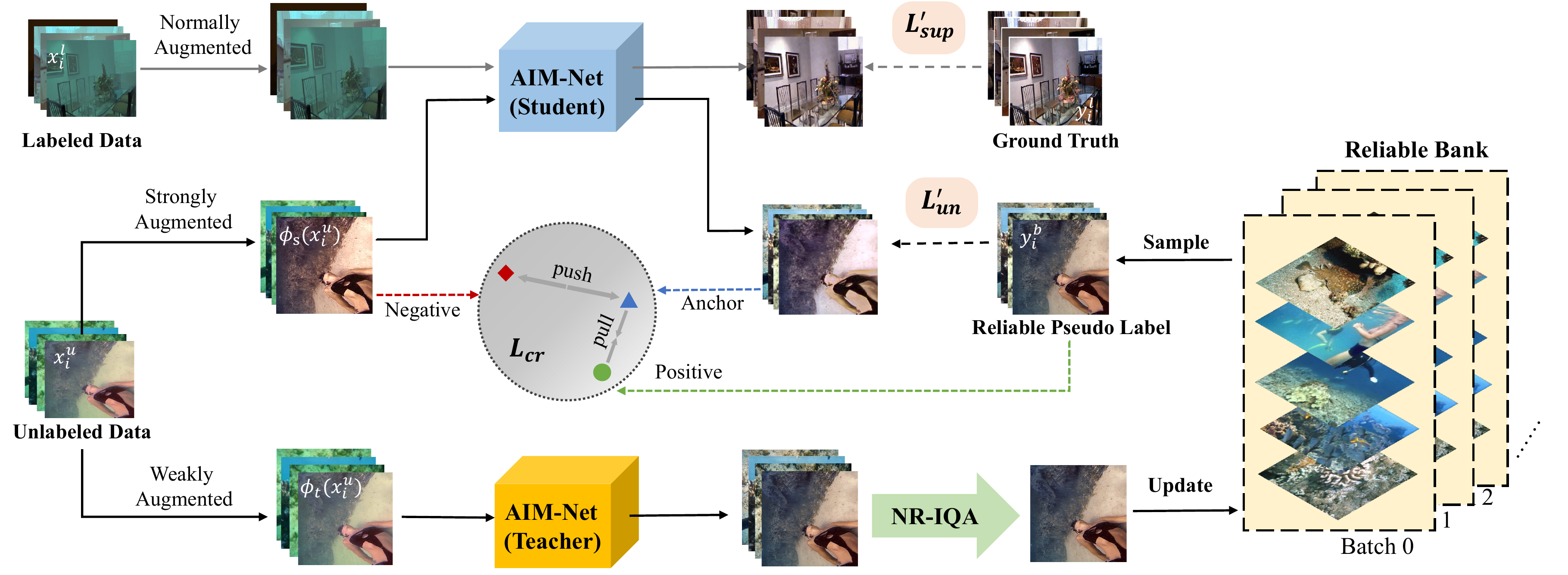}
\vspace{-0.3cm}
\caption{Illustration of our framework Semi-UIR. Semi-UIR is based on the mean teacher scheme with a student model and a teacher model. To guarantee the reliability of pseudo labels for unlabeled data, we build a reliable bank to archive best-ever teacher outputs measured by NR-IQA. Reliable pseudo labels guide the student's training via the unsupervised teacher-student consistency loss $L_{un}'$ and contrastive loss $L_{cr}$. The weights of the student are updated by minimizing the supervised loss ($L_{sup}'$) and unsupervised losses ($L_{un}'$ and $L_{cr}$). The teacher is updated with EMA from the student. 
\label{framework}}
\end{figure*} 

\subsection{Contrastive Learning}
Contrastive learning has emerged as an effective  paradigm in  self-supervised learning \cite{chen2020simclr, he2020moco, grill2020byol}. It enables  visual representation to be learned empirically by instance discrimination, through which similar samples are kept close to each other whereas dissimilar samples are separated far apart.
To take advantage of contrastive learning in image restoration, previous works largely focus on the construction of contrastive samples and feature space. For example, \cite{wu2021contrastive, contrastive_deblurring} 
take clean images as positive instances and degraded images as negative instances and then project them into a new feature space by VGG\cite{vgg}. Note that in the above works, contrastive loss is applied in a supervised manner, which is infeasible for unlabeled data.  How to make contrastive loss applicable to unlabeled data remains an open problem. \cite{cwr} is the first to employ contrastive learning in the context of  underwater image restoration. 
However, it is still a supervised learning method in nature, and contrastive loss is used as a regularization term to boost the performance of supervised learning. Against this backdrop, the present work puts forward a systematic approach for making use of  contrastive learning to exploit unlabeled data. 
\section{Method}
\subsection{Problem Formulation}
Semi-supervised learning aims to enable a learning system to learn from both labeled and unlabeled data. The problem for underwater image restoration is defined as follows. Let $D_L=\{(x_i^l,y_i^l)|x_i^l\in \mathcal{I}_s^{LQ},y_i^l\in \mathcal{I}_s^{HQ}\}_{i=1}^N$ denote the labeled dataset, where $x_i^l$ and $y_i^l$ are respectively the underwater image and clean ground truth from degraded set $\mathcal{I}_s^{LQ}$ and ground truth set $\mathcal{I}_s^{HQ}$. Similarly, let $D_U=\{x_i^u|x_i^u\in \mathcal{I}_u^{LQ}\}_{i=1}^M$ denote the unlabeled dataset, where $x_i^u$ is the underwater image sampled from the degraded set $\mathcal{I}_u^{LQ}$. It is worth mentioning that the data in $D_L$ and $D_U$ are disjoint, \ie $D_L \cap D_U = \emptyset$. Our goal is to learn a mapping on  $D = D_L \cup D_U$ that converts an underwater image $x$ to its clean counterpart $y$.

\subsection{Semi-supervised Underwater Restoration}
Our semi-supervised learning framework follows the typical setup in semi-supervised learning\cite{tarvainen2017mean, sohn2020fixmatch}, as illustrated in Fig. \ref{framework}. Specifically, our Semi-UIR consists of two networks of the same structure, called teacher and student respectively. The two networks differ mainly in how their weights are updated. 

The teacher's weights 
$\theta_t$ are updated by exponential moving average (EMA) of the student's weights $\theta_s$:
\begin{equation}\label{eq1}
    \theta_t=\eta \theta_t + (1-\eta) \theta_s,
\end{equation}
where $\eta\in(0,1)$ is the momentum. Using this update strategy, the teacher model can aggregate previously learned weights immediately after each training step.  As is noted in \cite{polyak1992acceleration}, temporal weight averaging  can stabilize the training process and help improve the performance compared with standard gradient descent. 

The weights of student network $\theta_s$ are updated using  gradient descent. Usually, the optimization of the student network can be formulated as minimizing the following loss:
\begin{equation} \label{eq:loss_sec3.2}
\centering
\begin{aligned}
    L_{total} = L_{sup} + \eta L_{un},
\end{aligned}
\end{equation}

\noindent
where $L_{sup} = \sum_{i=0}^N|f_{\theta_s}(x_i^l)-y_i^l|$ denotes the supervised loss and $L_{un}= \sum_{i=0}^M|f_{\theta_s}(\phi_s(x_i^u))-f_{\theta_t}(\phi_t(x_i^u))|$ represents the unsupervised teacher-student consistency loss. $|\cdot|$ refers to L1 distance. $\phi_s$ and $\phi_t$ are respectively the data augmentations of student's inputs and teacher's inputs.

Ideally, since the teacher network is in general better than the student network, $L_{un}$ could provide effective supervision to train the student network on the unlabeled dataset. We thus refer to the teacher's output $\hat{y}_i^u = f_{\theta_t}(\phi_t(x_i^u))$ as pseudo label. However, it is not guaranteed that the outputs of the teacher  are consistently better than those of the student. Wrong pseudo labels can potentially 
jeopardize the training of the student network.

\subsection{Reliable Teacher-Student Consistency}
To address the above issue, we shall select the reliable outputs of teacher as pseudo labels.
In image classification and semantic segmentation \cite{sohn2020fixmatch, nips2021ael, cvpr2022u2pl}, the reliability of the network's outputs is usually measured by entropy and confidence. However, the extension to image restoration problems does not appear to be straightforward due to the presence of new challenges. 
In particular, as a regression task, underwater image restoration  requires recovering  fine textures and removing color cast.  

To this end, we propose a reliable bank to store the best-ever outputs of the teacher network during the training process. To be specific, We first initialize our reliable bank to be an empty set, \ie $\mathcal{B}_U = \emptyset$.
In each training iteration, we compare the current output of teacher with both the student's output and the pseudo label in the reliable bank. If the teacher's output is the best in quality, then we replace the pseudo label in the reliable bank with the teacher's current  output. In this way, we could maintain a reliable bank $\mathcal{B}_U = \{y_i^b\}_{i=1}^M$. Note that $D' = D_U \cup \mathcal{B}_U = \{(x_i^u, y_i^b)\}_{i=1}^M$ is a pseudo labeled dataset.
This reliable bank can keep track of the best pseudo labels and therefore avoid the wrong labels involved in the calculation of the unsupervised consistency loss $L_{un}$. Then, we can re-write the $L_{un}$ in Eq. (\ref{eq:loss_sec3.2}) as:

\begin{equation}\label{eq:un_sec3.3}
    L_{un}' = \sum_{i=0}^M|f_{\theta_s}(\phi_s(x_i^u))-y_i^b|.
\end{equation}

Now arises the obvious question: How to determine the quality of a prediction without the true label? 

\subsection{Reliable Metric Selection} \label{sec:reliable_selection}
Intuitively, we could resort to non-reference image quality assessment (NR-IQA). Unfortunately, as is noted in \cite{berman2020underwater, UIEBD}, the commonly used UCIQE\cite{UCIQE} and UIQM\cite{UIQM} cannot accurately reflect the quality of restored underwater images. Therefore,  building our reliable bank based on such metrics is questionable. To find the best possible NR-IQA for underwater images, we conduct an empirical analysis of several NR-IQA metrics. 

\begin{figure}[!t]
\centering
\includegraphics[width=1.0\linewidth]{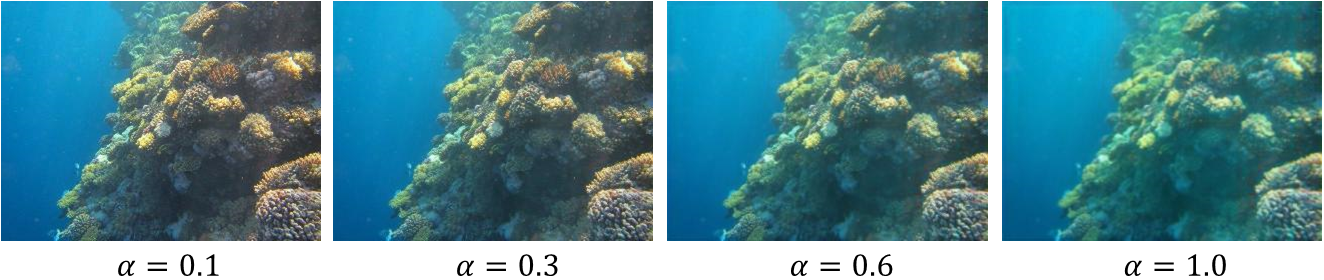}
\vspace{-0.6cm}
\caption{Examples of image fusion based on different $\alpha$.\label{sys_alpha}}
\end{figure}

\begin{figure}[!t]
\centering
\includegraphics[width=0.9\linewidth]{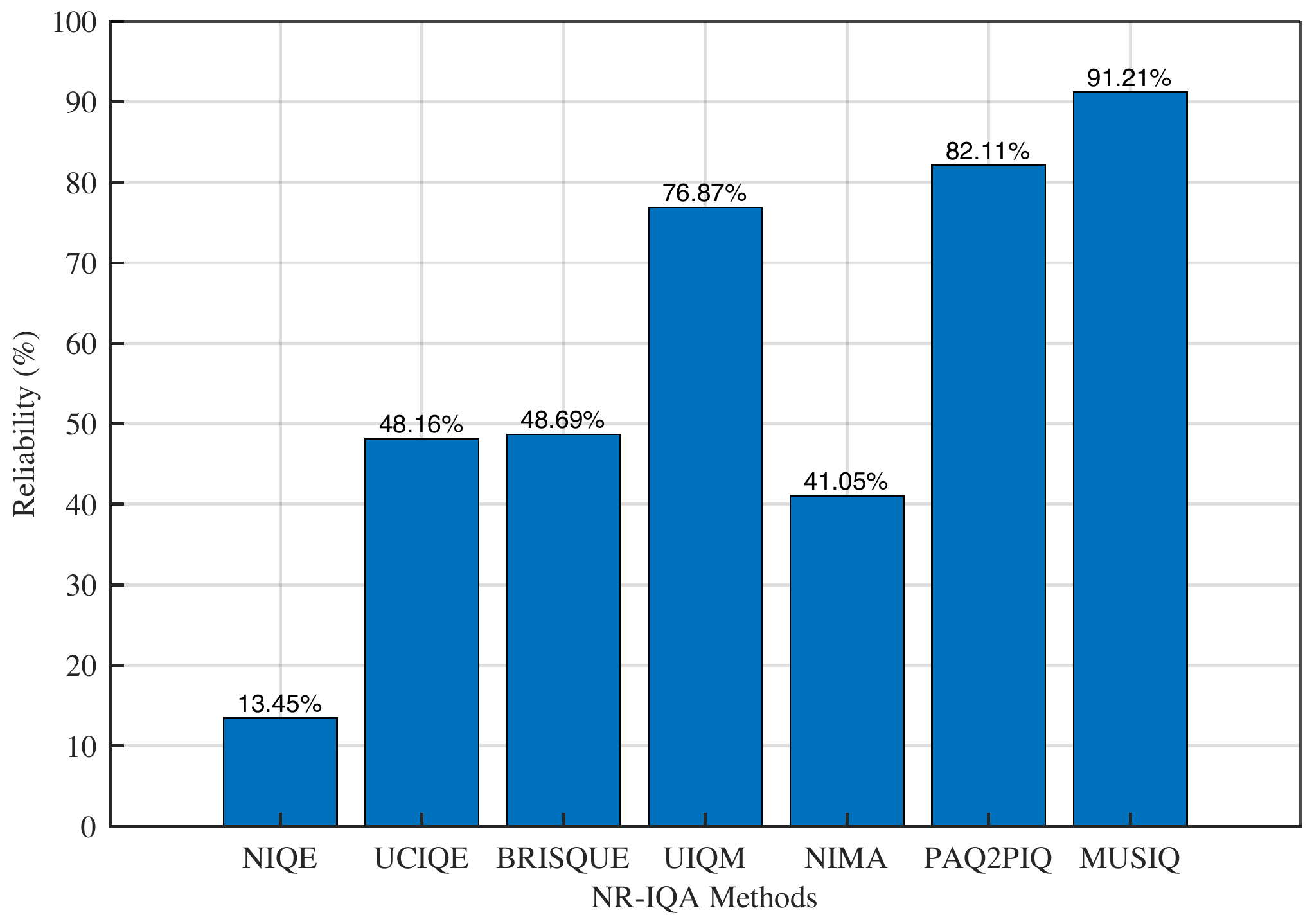}
\vspace{-0.4cm}
\caption{The results of different non-reference IQA indicators on EUVP benchmark, including UIQM\cite{UIQM}, UCIQE\cite{UCIQE}, BRISQUE\cite{BRISQUE}, NIQE\cite{NIQE}, NIMA\cite{talebi2018nima}, PAQ2PIQ\cite{PAQ2PIQ} and MUSIQ\cite{ke2021musiq}.}
\vspace{-0.3cm}
\label{nr}
\end{figure}

Given a degraded underwater image $x^l$ and a paired clean image $y^l$, we perform various linear combination of them to get a set of images with different quality. Specifically, let $\alpha_i = 0.1 \times i, i=1,2,...,10$, we can obtain a set of ten images $ \{\alpha_i x^l + (1-\alpha_i )y^l\}_{i=1}^{10}$. 
With the increase of $\alpha_i$, the visual quality of the corresponding image deteriorates, as shown in Fig. \ref{sys_alpha}. It thus makes sense to evaluate the NR-IQA metrics based on how well they capture this monotonicity law. In particular, an NR-IQA metric is identified as reliable if its score on the $ \alpha_i x^l + (1-\alpha_i )y^l$ decreases with the increase of $\alpha_i$.  Following this rule, we conduct experiments with seven NR-IQA approaches on EUVP benchmark\cite{EUVP-FUnieGan}, as it covers a wide range of underwater scenes. The experimental results are shown in Fig. \ref{nr}. 
We can observe that the deep learning based MUSIQ\cite{ke2021musiq} is most in line with the monotonicity law. Therefore, it is selected to measure the reliability of the networks' outputs.

The overall procedure of our reliable bank construction is summarized in Algorithm \ref{al:alg}.

\begin{algorithm}[!t]
\caption{Update of Reliable Bank}
\label{al:alg}
\begin{algorithmic}
\STATE \textbf{Require:} NR-IQA method $\Psi(\cdot)$;
\STATE Initialize $\mathcal{B}_U = \emptyset$;
\STATE Sample a batch of unlabeled images $\{x_i^u\}_{i=1}^b$ from $D_U$;
\FOR{each $x_i^u$}
\STATE Get teacher's prediction: $\hat{y}_i^u = f_{\theta_t}(\phi_t(x_i^u))$;
\STATE Get student prediction:
$\tilde{y}_i^u= f_{\theta_s}(\phi_s(x_i^u))$;
\STATE Compute NR-IQA scores of $\hat{y}_i^u$, $\tilde{y}_i^u$ and $y_i^b\in \mathcal{B}_U$:\\
$z_t = \Psi(\hat{y}_i^u)$, $z_s = \Psi(\tilde{y}_i^u)$, $z_b = \Psi(y_i^b)$;
\IF{$z_t > z_s$ and $z_t > z_p$}
\STATE Replace the $y_i^b$ in $\mathcal{B}_U$ by $\hat{y}_i^u$;
\ENDIF
\ENDFOR

\end{algorithmic}
\end{algorithm}

\subsection{Contrastive Regularization}
Typically, many mean teacher based methods uses L1 distance as consistency loss as is shown by $L_{un}$ in Eq. (\ref{eq:loss_sec3.2}). The simple consistency loss can easily make the student model overfit on wrong predictions, resulting in confirmation bias. To address this problem, we introduce contrastive loss in the training.
Contrastive learning has emerged as an effective paradigm in the self-supervised domain \cite{chen2020simclr, he2020moco, grill2020byol}. The goal of contrastive learning is to enable a model to produce a similar representation of \textit{positive} pairs and a dissimilar representation of \textit{negative} pairs. 
Recently, it has been extended to address image restoration problems \cite{wu2021contrastive, contrastive_deblurring}. Despite the tremendous success, they usually construct contrastive loss on paired datasets, where the \textit{positive} and \textit{negative} samples are respectively the labels and degraded images. In this section, we propose to incorporate contrastive loss in handling unlabeled data. To achieve this, we first need to construct positive and negative pairs. \cite{contrastive_super} provide an idea of directly using the teacher's output as positive samples. However, due to the wrong label problem we have discussed in the previous sections, using the teacher's outputs as positive samples might be harmful.

Thanks to our proposed reliable bank, where the samples are potentially of higher quality than the student's outputs, we can take $y_i^b$ as our positive sample. 
For the negative sample, we follow \cite{wu2021contrastive,contrastive_deblurring,contrastive_super} to take the strongly augmented degraded image $\phi_s(x_i^u)$ as our negative sample. 
After constructing the positive and negative samples, we can calculate the contrastive loss as follows:
\begin{equation}\label{eq4}
L_{cr}=\sum_{j=1}^K \sum_{i=1}^M \omega_j\frac{|\varphi_j(\tilde{y}_i^u),\varphi_j(y_i^b)|}{|\varphi_j(\tilde{y}_i^u),\varphi_j(\phi_s(x_i^u))|},
\end{equation}
\noindent
where $\tilde{y}_i^u= f_{\theta_s}(\phi_s(x_i^u))$ is the student's prediction on the unlabeled dataset $D_U$. $\varphi_j(\cdot)$ represents the $j_{th}$ hidden layer of the pre-trained VGG-19\cite{vgg} and $\omega_j$ is the weight coefficient. We use L1 loss to measure the distance in feature space between the students' outputs with the positive and negative samples. 

\subsection{Overall Optimization Objective}
Similar to Eq. (\ref{eq:loss_sec3.2}), our final optimization objective consists of supervised loss and unsupervised loss.

For the supervised loss, unlike the one defined in Eq. (\ref{eq:loss_sec3.2}) that only calculate the L1 distance, we follow \cite{prwnet} to extend the original $L_{sup}$ by adding perceptual loss $L_{per}$ and gradient penalty $L_{grad}$:
% $L_{sup}' = L_{sup}+ \beta_1 L_{per}+\beta_2 L_{grad}$.
\begin{equation}\label{eq:loss_sup_all}
    L_{sup}' = L_{sup}+ \beta_1 L_{per}+\beta_2 L_{grad}.
\end{equation}
% \noindent 

For the unsupervised loss, we replace the original $L_{un}$ by a combination of the proposed reliable teacher-student consistency loss and contrastive loss: 
% $L_{un}'' = L_{un}'+ L_{cr}$.
\begin{equation}\label{eq:loss_un_all}
    L_{un}'' = L_{un}'+ \gamma L_{cr}.
\end{equation}

Finally, we rewrite our overall optimization objective following Eq. (\ref{eq:loss_sec3.2}):

\begin{equation}\label{eq:loss_all}
    L_{overall} = L_{sup}'+ \lambda L_{un}''.
\end{equation}

 Due to the page limit, please refer to supplementary material for the detailed perceptual loss and gradient penalty.

\section{Experimental Results}

% {\color{red}
\subsection{Implementation Details}
\begin{figure}[t]
\includegraphics[width=1.0\linewidth]{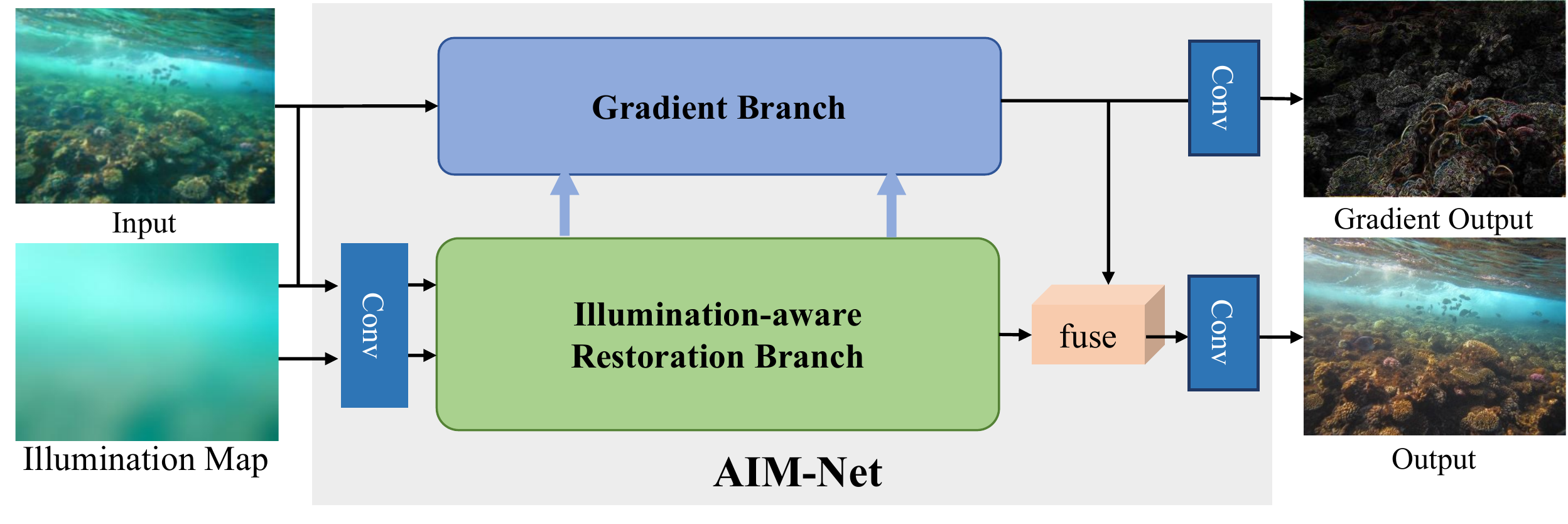}
\vspace{-0.8cm}
\caption{An overview of the proposed Asymmetric Illumination-aware Multi-scale Network (AIM-Net).}
\label{network}
\end{figure} 
\noindent
\textbf{Network Structure}
Our student and teacher model are based on the same structure, AIM-Net. To tackle the prominent issues with underwater images (e.g., low contrast, color distortion and blur), certain prior information of such images (e.g., illumination prior \cite{UWNR} and gradient prior) are effectively exploited.
As is shown in Fig. \ref{network}, the network consists of two branches: illumination-aware restoration branch and gradient branch. The restoration branch incorporates illumination prior to enhance the color and light source perception capabilities.  
The gradient branch is introduced to enhance the edge structure.
Please refer to supplementary material for detailed network structure.
\begin{table}[!t]
\centering
\small
\caption{Evaluations of different methods on full-reference benchmarks in terms of PSNR and SSIM. Best results are in \textbf{bold} and the second best results are with \underline{underline}.}
\scalebox{0.88}{
\setlength{\tabcolsep}{3mm}{
\begin{tabular}{@{}|c|c|c|c|c|@{}}
\toprule
\multicolumn{1}{|c|}{\multirow{2}{*}{\textbf{Method}}} & \multicolumn{2}{|c|}{\cellcolor[HTML]{FFCCC9}\textbf{testS}} &  \multicolumn{2}{|c|}{\cellcolor[HTML]{9AFF99}\textbf{testR}} \\ 
% \midrule
& \textbf{PSNR}$\uparrow$ & \textbf{SSIM}$\uparrow$ &  \textbf{PSNR}$\uparrow$ &\textbf{SSIM}$\uparrow$ 
\\
\midrule
Input &14.64 &0.641 &18.23 &0.746  \\
GDCP\cite{peng2018gdcp} &12.89 &0.576  &15.78 &0.757 \\
MMLE\cite{MMLE} &12.76 &0.651  &20.01 &0.781 \\
WaterNet\cite{UIEBD} &15.44 &0.706  &21.58 &0.858  \\
Ucolor\cite{ucolor} &\underline{23.32}  &\textbf{0.853}  &\underline{22.92}&\underline{0.881}  \\
PRWNet\cite{prwnet}  &17.27 &0.723 &20.98 &0.848 \\
FGAN\cite{EUVP-FUnieGan}  &18.54 &0.743 &19.41 &0.824  \\
CWR\cite{cwr}  &14.79  &0.697 &21.87 &0.815 \\
{\bf Semi-UIR}  &\textbf{23.40} &\underline{0.821}  &\textbf{24.59} &\textbf{0.901} \\
\bottomrule
\end{tabular}}}
\vspace{-0.2cm}
\label{tab:paired}
\end{table}

\begin{figure*}[!t]
\centering
\includegraphics[width= \linewidth]{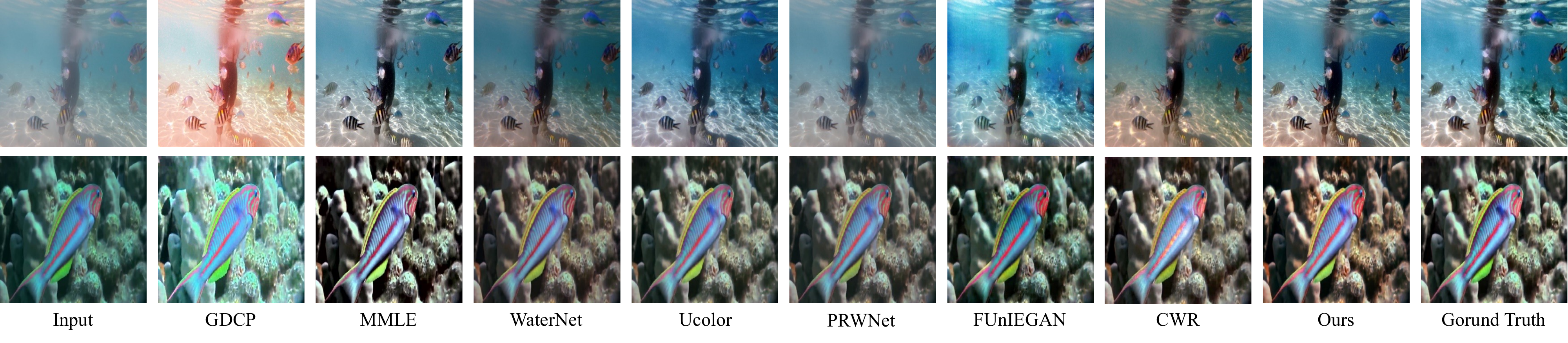}
\vspace{-0.8cm}
\caption{Visual comparisons of full-reference data from UIEB benchmark.\label{fig:experiments_testR}}
\end{figure*}  

% \subsubsection{Training Details}
\noindent
\textbf{Training Details}
Our method is implemented using Pytorch library \cite{paszke2019pytorch} and conducted on NVIDIA RTX 3090 GPUs.
We use AdamP\cite{heo2021adamp} as our optimizer. In consideration of its fast convergence to optimum, we select AdamP mainly to reduce the training time. During training, we use a mini-batch size of 16, where 8 samples are labeled and 8 samples are unlabeled. The initial learning rate is set to $2e^{-4}$. We train for 200 epochs with the learning rate multiplied by 0.1 at 100 epochs. The training images are all cropped to a size of $256 \times 256$.
For the data augmentation on unlabeled data, we only apply resize on the teacher's inputs and impose strong data augmentation on the student's inputs. The strong augmentation includes resize, color jitter, gaussian blur and gray scale. Labeled data is normally augmented, including resize, random crop and rotation.
The weights of different loss components are set as follows: $\beta_1 = 0.3$, $\beta_2 = 0.1$, $\gamma = 1$ and $\lambda$ is updated with training epoch $t$ following an exponential warming up function \cite{acmm2021dmt}: $\lambda (t)=0.2\times e^{-5(1-t/200)^2}$. 

\subsection{Datasets}
Our training set contains 1600 labeled image pairs and 1600 unlabeled images. The labeled image pairs are randomly sampled from \cite{Anwar2019UWCNN} and UIEB\cite{UIEBD} with a ratio of 1:1. \cite{Anwar2019UWCNN} provides a synthesized underwater image dataset in indoor scene.
The UIEB\cite{UIEBD} dataset contains 890 real underwater images with corresponding ground truths. 
The unlabeled images are sampled from the unpaired data in the EUVP benchmark\cite{EUVP-FUnieGan}, which cover a variety of underwater scenes, water types and lighting conditions. 

Test set is built with full-reference and non-reference benchmarks. Full-reference test set includes 110 pairs from \cite{Anwar2019UWCNN} and 90 pairs from UIEB\cite{UIEBD}, namely testS and testR. Non-reference test set includes nearly 700 real world underwater images without ground truths from benchmarks such as UIEB, EUVP, RUIE\cite{liu2020real} and Seathru\cite{seathru}.

\subsection{Comparison with the State-of-the-Arts}
We compare our proposed Semi-UIR with seven state-of-the-art underwater restoration methods, including two traditional methods (GDCP\cite{peng2018gdcp}, MMLE\cite{MMLE}) and five deep learning based methods (WaterNet\cite{UIEBD}, Ucolor\cite{ucolor}, FUnIE-GAN\cite{EUVP-FUnieGan}, PRWNet\cite{prwnet} and CWR\cite{cwr}). All the compared methods are re-trained on our training dataset.
For testS and testR, we conduct full-reference evaluations using PSNR, SSIM\cite{ssim}.
For non-reference test set, we provide evaluation results using UIQM\cite{UIQM}, UCIQE\cite{UCIQE} and MUSIQ\cite{ke2021musiq}.

\noindent
\textbf{Results on full-reference datasets.} The quantitative results on testS and testR are shown in Table \ref{tab:paired}. On testS, our method performs the best in PSNR, but slightly worse than Ucolor in terms of SSIM. One potential reason is that incorporating unlabeled real underwater images in training might emphasize the network to pay more attention to the real underwater scene. This can be confirmed by checking the quantitative results on testR. On testR, our method outperforms the other methods by a significant margin (outperforms the second best by 1.67dB in PSNR).  
In addition to the quantitative results, qualitative results are shown in Fig. \ref{fig:experiments_testR}. Our results are visually pleasant, while the compared methods suffer from color cast and over-enhancement. 

\begin{table*}[!t]
\centering
\small
\caption{Evaluations of different methods on non-reference benchmarks in terms of UIQM, UCIQE and MUSIQ. Best results are in \textbf{bold} and the second best results are with \underline{underline}.}
\scalebox{0.91}{
\setlength{\tabcolsep}{2.4mm}{
\begin{tabular}{@{}|c|c|c|c|c|c|c|c|c|c|c|c|c|@{}}
\toprule
\multicolumn{1}{|c|}{\multirow{2}{*}{\textbf{Method}}} & \multicolumn{4}{|c|}{\cellcolor[HTML]{FFCCC9}\textbf{UIQM} (higher, better)} &  \multicolumn{4}{|c|}{\cellcolor[HTML]{9AFF99}\textbf{UCIQE} (higher, better)}&\multicolumn{4}{|c|}{\cellcolor[HTML]{5DADE2}\textbf{MUSIQ} (higher, better)}\\ 
% \midrule
& \textbf{UIEB} & \textbf{EUVP} & \textbf{RUIE} &\textbf{Seathru} &\textbf{UIEB} & \textbf{EUVP} &\textbf{RUIE} & \textbf{Seathru}&\textbf{UIEB} & \textbf{EUVP} &\textbf{RUIE} & \textbf{Seathru}\\
\midrule
Input &3.066 &4.729 &3.948 &5.925 &0.509 &0.517 &0.490 &0.537&41.70&42.73&33.53&60.25\\
GDCP\cite{peng2018gdcp} &3.401 &4.738 &4.509 &5.343 &0.564 &\textbf{0.599} &0.565 &0.590 &40.07 &42.49 &34.63 &60.54\\ 
MMLE\cite{MMLE} &4.283 &4.723 &\textbf{4.967} &5.555 &\underline{0.578} &\underline{0.596} &\underline{0.571} &0.620&\underline{40.33}&\underline{47.55}&\underline{36.80}&\underline{66.16}\\
WaterNet\cite{UIEBD} &4.118 &\underline{5.317} &4.568 &\underline{6.829} &0.572 &0.595 &\textbf{0.572} &0.610 &40.32&43.07&32.23&64.38\\
Ucolor\cite{ucolor} &3.894 &5.286 &4.426 &6.752 &0.542 &0.566 &0.534 &0.594&40.08&41.81&33.66&64.44\\
PRWNet\cite{prwnet}  & \underline{4.371} &\textbf{5.330} &4.395 &6.778 &0.518 &0.543 &0.518 &0.572&40.30&43.52&33.12&62.82\\
FGAN\cite{EUVP-FUnieGan}  & 4.315 &4.469 &4.519 &4.853 &0.541 &0.561 &0.527 &0.564&40.95&43.36&34.48&64.25 \\
CWR\cite{cwr}  & 4.133 &5.152 &4.469 &6.067 & \textbf{0.587} & \underline{0.596} &0.565 &\underline{0.624}&38.46&41.46&31.25&64.21 \\
{\bf Semi-UIR}  &\textbf{4.598} &5.291 &\underline{4.671} & \textbf{6.846} & \textbf{0.587} &0.593 &0.557 &\textbf{0.632} &\textbf{43.77}&\textbf{51.66}&\textbf{37.87}&\textbf{66.61}\\
\bottomrule
\end{tabular}}}
\label{tab:unpaired}
\end{table*}

\begin{figure*}[!t]
\centering
\includegraphics[width= 0.97\linewidth]{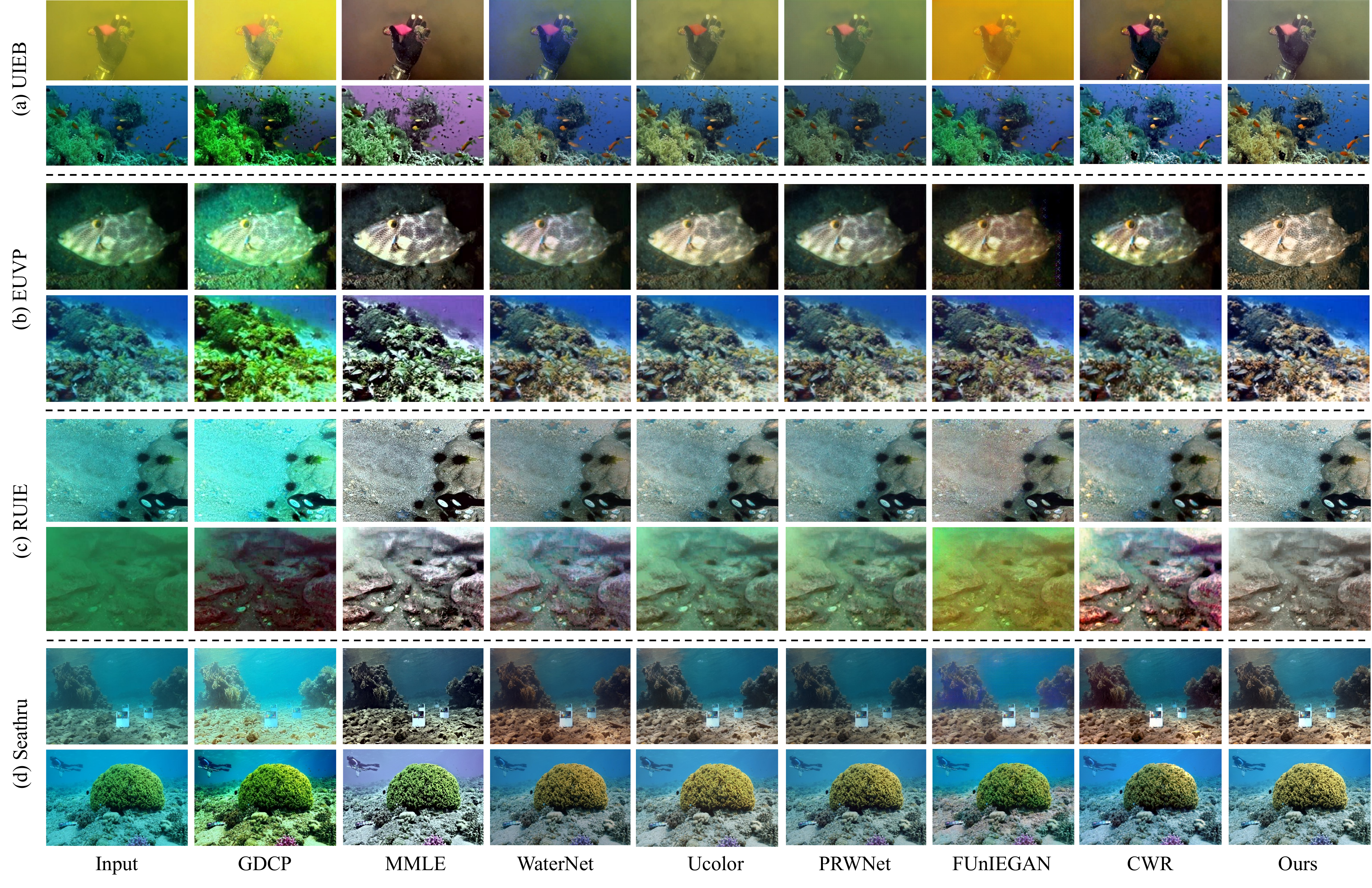}
\vspace{-0.4cm}
\caption{Visual comparisons on non-reference benchmarks UIEB\cite{UIEBD}, EUVP\cite{EUVP-FUnieGan}, RUIE\cite{liu2020real} and Seathru\cite{seathru}}
\label{fig:experiments_testN1}
\end{figure*}  

\noindent
\textbf{Results on non-reference datasets.}
The quantitative results on UIEB, EUVP, RUIE and Seathru are shown in Table \ref{tab:unpaired}. By quickly checking throughout the table, we can observed that our method significantly outperform the compared method in MUSIQ. Besides, we also achieve
competitive performance in terms of UIQM and UCIQE. However, as is noted in\cite{berman2020underwater, UIEBD, guo2023uranker}, UIQM and UCIQE  might be biased to some characteristics and thus cannot accurately reflect the true visual quality of restored images. Similarly, the performance in MUSIQ is also for a reference.
Therefore, the quantitative results might be insufficient to indicate the quality of restored underwater images due the  underdeveloped NR-IQA metrics. We further show the qualitative results on the four benchmarks in Fig. \ref{fig:experiments_testN1}. Compared with other methods, our approach can robustly restore various types of underwater images with natural color and rich details. Under the guidance of reliable pseudo labels and powerful contrastive regularization, our framework generalizes well on various underwater scenes.

\subsection{Ablation Study}

To analyze the effectiveness of Semi-UIR, we conduct ablation studies to reveal the influence of the key components in our method. They are presented as follows: (a) \textbf{Sup-base}: where we train the network AIM-Net without semi-supervised learning and unlabeled data.
(b) \textbf{Semi-base}: Base semi-supervised training with consistency loss $L_{un}$.
(c) \textbf{Semi-base+RB*}: using reliable bank based on \textbf{Semi-base} without contrastive loss.
(d) \textbf{Semi-base+CL*}: adding contrastive loss to \textbf{Semi-base}, without using reliable bank.
(e) \textbf{Semi-UIR}: our proposed Semi-UIR.

The quantitative results of above methods are shown in Table \ref{tab:ablation}. We can observe that our full solution performs best. In addition, by comparing \textbf{Semi-base+RB*} with \textbf{Semi-base} and Semi-UIR with \textbf{Semi-base+CL*}, it is easy to verify the effectiveness of incorporating the reliable bank. 

\begin{figure}[]
\includegraphics[width=1.0\linewidth]{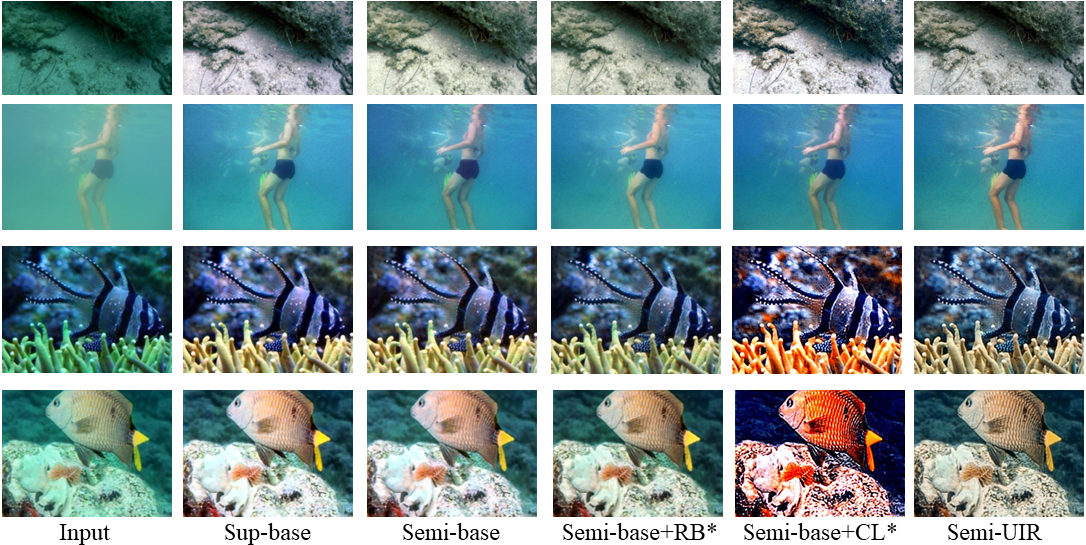}
\vspace{-0.6cm}
\caption{Results of ablation study about Semi-UIR.}
\label{fig:ablation}
\end{figure}

Besides, the qualitative results are shown in Fig. \ref{fig:ablation}, where special attention should be paid on \textbf{Semi-base+CL*} \textbf{Semi-base+RB*}. (1) In \textbf{Semi-base+CL*}, without the reliable positive samples, the contrastive loss pushes the network to produce extremely different results than the negative samples (inputs). However, this unfortunately results in over-enhancement. (2) On the contrary, in \textbf{Semi-base+RB*}, without the help of contrastive loss, the restored images still suffer from color distortion and are close to the degraded inputs. The two ablation studies verify the utility of the reliable bank and contrastive regularization.

 \begin{table}[!t]
\centering
\small
\caption{Ablation studies on testR, EUVP and UIEB benchmarks in terms of PSNR or MUSIQ. MT denotes mean teacher framework, CL represents contrastive loss, and RB is reliable bank.}
\vspace{-0.3cm}
\scalebox{0.9}{
\setlength{\tabcolsep}{1.8mm}{
\begin{tabular}{|c|c|c|c|c|c|c|c|c|}
\toprule
\textbf{Method} &\textbf{MT} &\textbf{RB} &\textbf{CL} & \textbf{testR} & \textbf{EUVP} &\textbf{UIEB}\\ 
% \midrule
\midrule
\textbf{Sup-base} & & & &24.38 &42.66 &40.70\\
\textbf{Semi-base} &$\surd$ & & &23.11 &42.48  &40.33\\
\textbf{Semi-base+RB*} &$\surd$ &$\surd$ & &24.53 &43.27 &42.16\\
\textbf{Semi-base+CL*} &$\surd$ & &$\surd$ &23.97 &46.59  &40.64\\
{\bf Semi-UIR} &$\surd$ &$\surd$ &$\surd$ &\textbf{24.59} &\textbf{51.66} &\textbf{43.77}\\
\bottomrule
\end{tabular}}}
\vspace{-0.2cm}
\label{tab:ablation}
\end{table}

\subsection{Breakdown of the Training}
\begin{figure}[h]
\centering
\includegraphics[width=1.0\linewidth]{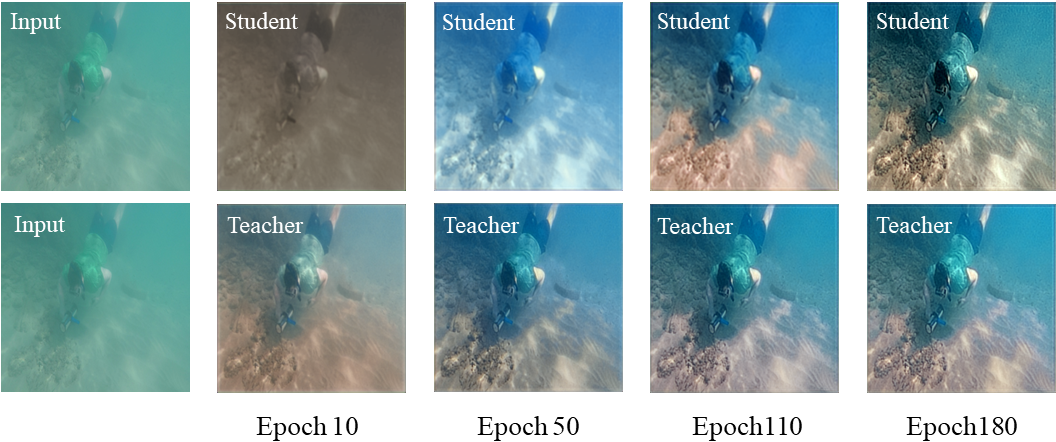}
\vspace{-0.6cm}
\caption{Examples of intermediate predictions of the teacher model and student model.
}
\vspace{-0.3cm}
\label{mid-outputs}
\end{figure}

To further illustrate that the teacher's outputs can be used to train student network,
we here provide some intermediate results during training. The results are shown in Fig. \ref{mid-outputs}. At the beginning of the training (10 epochs), the teacher's prediction is much better than that of student. As the training processes, the student's outputs and teacher's outputs are improved simultaneously.

\subsection{Influence of Non-reference Metric}
We here conduct experiments to show the influence of using different NR-IQA approaches in building our reliable bank. We conclude in Sec. \ref{sec:reliable_selection} that MUSIQ is the most reliable one. To further demonstrate the correctness of this selection, we here show the final performance of using NIMA, PAQ2PIQ and MUSIQ on the labeled dataset, \ie testS and testR. Table \ref{tab:NR-IQA} shows the results. It can be observed that we can achieve the best performance by using MUSIQ. It also shows that using PAQ2PIQ is better than using NIMA, which is consistent with their reliability shown in Fig. \ref{nr}.
\begin{table}[h]
\centering
\small
\caption{Evaluation the influence of adopting different NR-IQA metrics on testS and testR.}
\vspace{-0.3cm}
\scalebox{0.9}{
\setlength{\tabcolsep}{2.6mm}{
\begin{tabular}{@{}|c|c|c|c|c|c|@{}}
\toprule
\multicolumn{1}{|c|}{\multirow{2}{*}{\textbf{Method}}} &
\multicolumn{1}{|c|}{\multirow{2}{*}{\textbf{Reliability}}} &
\multicolumn{2}{|c|}{\cellcolor[HTML]{FFCCC9}\textbf{PSNR}} &  \multicolumn{2}{|c|}{\cellcolor[HTML]{9AFF99}\textbf{SSIM}} \\ 
% \midrule
& &\textbf{testS} & \textbf{testR} & \textbf{testS} &\textbf{testR}\\
\midrule
% BRISQUE &23.15&24.00&0.820&0.900\\
NIMA &41.05\%&23.01 &23.88 &0.815 &0.888\\
PAQ2PIQ &\underline{82.11}\%&\underline{23.08} &\underline{24.28} &\underline{0.818} &\underline{0.893}\\
 MUSIQ &\textbf{91.21}\%&\textbf{23.40} &\textbf{24.59} &\textbf{0.821} &\textbf{0.901}\\
\bottomrule
\end{tabular}}}
\label{tab:NR-IQA}
\end{table}

\subsection{Influence of Data Augmentation}
 
We finally
show the influence of using different data augmentations in Table \ref{tab:data augmentation}. 
By comparing the method using data augmentation with baseline, it is easy to conclude that adopting any of the data augmentations is beneficial. Besides, using a mixture of the three strategies achieves the best performance.

\begin{table}[h]
\centering
\small
\caption{Evaluation of using different data augmentation. Baseline is our full solution without using the three strong data augmentations. Numbers are either in SSIM or MUSIQ.}
\vspace{-0.3cm}
\scalebox{0.92}{
\setlength{\tabcolsep}{2mm}{
\begin{tabular}{|c|c|c|c|c|c|}
\toprule
\textbf{Strategy} &\textbf{testR} &\textbf{UIEB} &\textbf{EUVP} &\textbf{RUIE} &\textbf{Seathru}\\ 
% \midrule
\midrule
\textbf{Baseline}&0.880 &40.12 &46.06 &31.14 &64.71 \\
\textbf{Color Jitter} &0.889 &40.31 &49.16 &33.66 &64.87 \\
\textbf{Gaussian Blur} &0.896 &41.23 &49.27 &36.88 &64.88 \\
\textbf{Gray Scale} &0.895 &40.61 &47.57 &32.51 &65.19 \\
\textbf{All} &\textbf{0.901} &\textbf{43.77} &\textbf{51.66} &\textbf{37.87} &\textbf{66.61} \\
\bottomrule
\end{tabular}}}
\vspace{-0.3cm}
\label{tab:data augmentation}
\end{table}

\section{Conclusion}
We propose an efficient semi-supervised underwater image restoration method named Semi-UIR.  As demonstrated the ablation experiments, the superior performance of the proposed method over other SOTA algorithms can be attributed to reliable teacher-student consistency and  contrastive regularization.  The follow-up research can be carried out in two directions: 1) extend the semi-supervised framework to cover other restoration tasks, 2) optimize memory usage during training and improve performance via  memory management.

%%%%%%%% REFERENCES
{\small
\bibliographystyle{ieee_fullname}
\bibliography{PaperForReview}
}

\appendix
% \onecolumn
\section{Supplementary Material}

In the supplementary material, we first introduce the overall structure of our AIM-Net, details of validation set and loss functions, and then analyze the influence of the optimizer and NR-IQA metrics during training. Meanwhile, inherent advantage of Semi-UIR and additional experimental results on non-reference benchmarks are provided as well.

\subsection{Details of the Network Structure}

\begin{figure*}[t]
\includegraphics[width=1.0\linewidth]{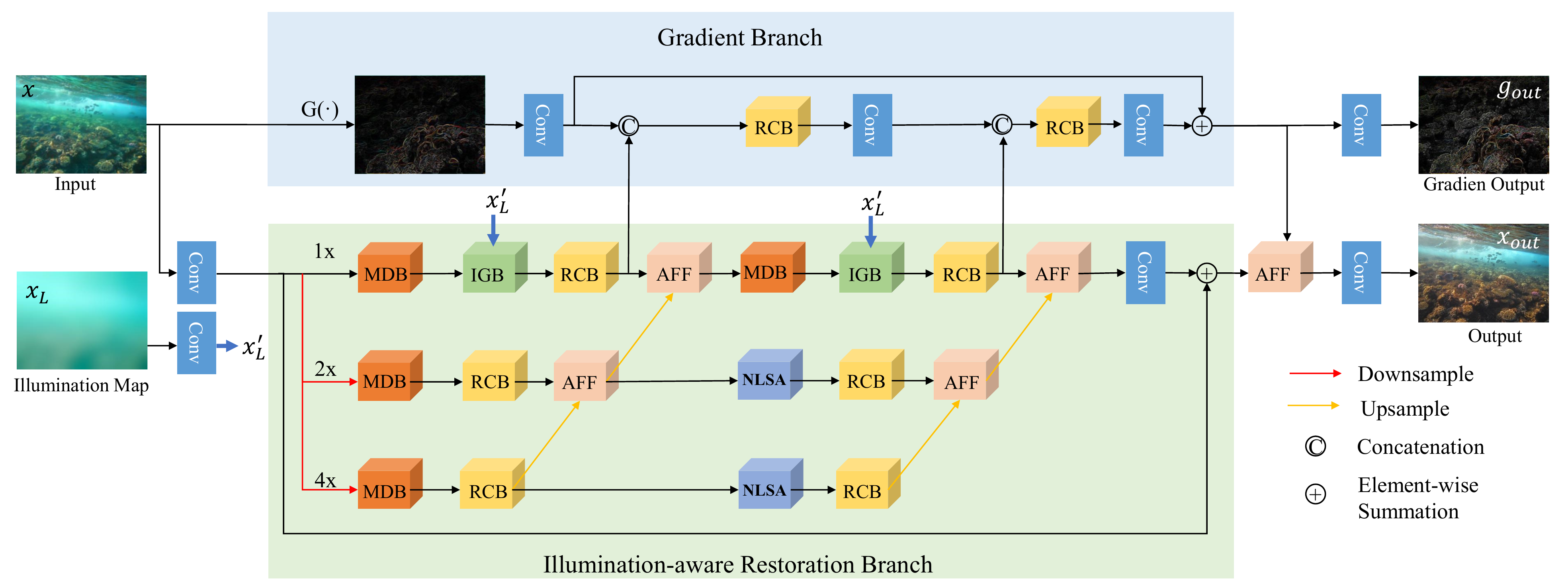}
\vspace{-0.5cm}
\caption{An overview of the proposed Asymmetric Illumination-aware Multi-scale Network (AIM-Net). AIM-Net mainly contains an illumination-aware restoration branch and a gradient branch.}
\vspace{-0.7cm}
\label{network1}
\end{figure*}

The goal of underwater image restoration is to conquer the problems such as low contrast, color distortion and blur. To this end, we propose an Asymmetric Illumination-aware Multi-scale Network (AIM-Net) to restore underwater images, as illustrated in Fig. \ref{network1}. AIM-Net contains two important branches, namely \textit{illumination-aware restoration branch} and \textit{gradient branch}. The restoration branch adopts a multi-scale structure containing three parallel streams to exact precise spatial information and abundant contextual information in different scales. In addition, illumination prior is incorporated into the restoration branch to enhance the perception capability of color and light source. The simple gradient branch aims to recover high quality gradient map. The recovered gradient features are merged into the restoration branch to provide edge and structure prior.

The entire pipeline is as follows. The degraded image $x$ passes through a convolution layer to extract preliminary features. After that, the preliminary features are fed into the illumination-aware restoration branch to extract and integrate precise spatial information and contextual information. The illumination map $x_L$ is converted to illumination feature $x_L'$ after a convolution layer. Then $x_L'$ participates in the restoration branch via illumination guidance block (IGB) \cite{AirNet}. On the other side, $x$ is fed into the gradient branch to restore fine gradient information. The gradient branch incorporates intermediate representations from the restoration branch because the intermediate representations carry rich structure information and are helpful for gradient map recovery. Later, attention feature fusion (AFF) block \cite{AFF} fuses the fine features obtained by the two branches. Finally, the restored clear image $x_{out}$ and restored gradient map $g_{out}$ are both obtained after a convolution layer.

\textbf{Illumination-aware Restoration Branch}
The task of the restoration branch is to reconstruct overall structure and color information of a degraded image. To complete the task, we introduce a parallel multi-scale structure incorporating illumination prior. The multi-scale structure can effectively integrate local details from different resolutions to maintain edge features and suppress halo artifacts \cite{wang2019underwater}. Therefore, we adopt the structure of Multiscale Residual Block (MRB) \cite{MIRNetv2} as the backbone. MRB contains three parallel streams of different scales. Each stream utilizes residual contextual block (RCB) to distill useful spatial information and selective kernel feature fusion (SKFF) block to integrate contextual information. Based on the structure of MRB, we add several fundamental modules to further enhance feature extraction capability, including: multi-dilated-convolution block (MDB), IGB, non-local spatial attention (NLSA) block. Meanwhile, we replace the SKFF block with AFF block to better aggregate contextual information. These modules are described in detail as follows.

MDB works at the front of each stream to exact spatial features. It expands the receptive field via combing four dilated convolutions. Dilation rates are set as 1, 2, 3 and 4, consistent with \cite{dilated_conv}. RCB \cite{MIRNetv2} distills useful contextual features by modeling and transforming the inter-channel dependencies via attention mechanism.

In the highest resolution stream, we introduce IGB between MDB and RCB to incorporate illumination prior. As is noted in \cite{UWNR}, the illumination map reflects underwater light field information including ambient light and scene-dependent degradation that are essential for underwater image formation. As a result, we introduce the illumination map to help the network perceive color and light source information.
The illumination map is estimated in accordance with \cite{UWNR}. IGB consists of a spatial feature transformation layer \cite{SFT} and a deformable convolution \cite{deformable}. In this way, our network can adapt to different color and light degradation types of underwater images.

To better integrate features from different scales, we replace SKFF module with AFF module. SKFF captures channel-wise dependencies via global-scale channel attention, whereas AFF squeezes local and global information into channel attention because local information is helpful to highlight local small targets \cite{AFF}. The AFF module follows the RCB to aggregate local and global contextual features. Furthermore, we add NLSA module \cite{NLSA} in the middle of the two lower resolution streams. NLSA models long-range feature correlations and enjoys robustness from sparse representation. Compared with the standard non-local attention \cite{NLA}, it reduces computation expenses significantly.

With the help of the above well-designed modules, the network is able to extract fine feature representations. These feature representations are important for texture and color reconstruction.

\textbf{Gradient Branch}
Image gradient map contains rich edge information and can guide the network to focus on local regions with sharp edges. As a result, we adopt the gradient map to enhance the edges of a restored image. Similar to \cite{spsr}, the image gradient prior is introduced to promote the restoration of underwater images from two aspects: 1) a gradient branch to restore a high-quality gradient map and provide structural information for the restoration branch; 2) a supplementary gradient loss to constrain the second-order relationship of adjacent pixels, guiding the underwater image restoration to focus more on the geometry. The gradient branch first estimates a coarse gradient map of $x$ via gradient operation $G(\cdot)$ \cite{spsr}, and then enhances the gradient map. Thanks to the intermediate representations from the restoration branch, the gradient branch can recover a fine gradient map with a very simple structure only including three convolution layers and two RCBs. Gradient loss is detailed in Eq. (\ref{eqnew}). Under these two types of guidance, the structure features can be better preserved, and the restoration results with sharper edges, higher perceptual quality and less geometric-inconsistent textures can be obtained.

In a word, AIM-net takes a degraded underwater image $x$ and its corresponding illumination map $x_L$ as input and outputs a restored clear image $x_{out}$ and a restored gradient map $g_{out}$:
\vspace{-0.2cm}
\begin{equation}\label{eq-net}
x_{out}, g_{out} = f_{\theta}(x, x_L),
\end{equation}
where $f_{\theta}$ represents the AIM-net parameterized by $\theta$. AIM-Net has 1.675M parameters, and its inference speed is 33.3 FPS on the images with a resolution of 256× 256.
\subsection{Details of Validation Set and Loss Functions}
Our validation set, independent of testS and testR, contains 200 pairs of full reference underwater images from \cite{Anwar2019UWCNN, UIEBD} with ratio 12:8.

The detailed perceptual loss and gradient penalty are as follows:
\begin{equation}\label{eqnew}
    \begin{aligned}
    L_{per}&=\sum_{j=1}^{K'}\sum_{i=1}^N|\varphi_j'(x_{out_i})-\varphi_j'(y_i^l)|\\
    L_{grad}&=\sum_{i=1}^N|g_{out_i}-G(y_i^l)|,
    \end{aligned}
\end{equation}
\noindent where $L_{per}$ denotes perpetual loss based on pretrained VGG-16\cite{vgg} network and $L_{grad}$ denotes gradient loss. $\varphi_j'$ refer to  ReLU1-2, ReLU2-2, and ReLU3-3 layers of the VGG-16 model. $y^l$ denotes clear ground truth. $G(\cdot)$ stands for the operation to extract a gradient map\cite{spsr}. $L_{grad}$ constrains the restored gradient map of AIM-Net to approach the ground truth's gradient map. 

\subsection{Influence of Optimizer and NR-IQA Metrics during Training}
Table \ref{tab:training optim} shows the total training epochs required by Adam and AdamP to reach a similar accuracy. It can be observed that the training time required by Adam is longer than that by AdamP. Thus we choose AdamP as training optimizer.
\vspace{-0.1cm}
\begin{table}[h]
\centering
\small
\caption{Evaluation the influence of adopting different optimizers on testR in terms of PSNR and SSIM.}
\vspace{-0.1cm}
\scalebox{1.0}{
\setlength{\tabcolsep}{2.4mm}{
\begin{tabular}{|c|c|c|c|}
\toprule
\textbf{Optim} &\textbf{PSNR} &\textbf{SSIM} &\textbf{Epochs}\\ 
% \midrule
\midrule
Adam&24.48&0.902 &240 \\
\textbf{AdamP} &24.59 &0.901 &200 \\
\bottomrule
\end{tabular}}}
\vspace{-0.3cm}
\label{tab:training optim}
\end{table}

\begin{figure}[h]
    \centering
    % \vspace{-0.4cm}
    \includegraphics[width=1.0\linewidth]{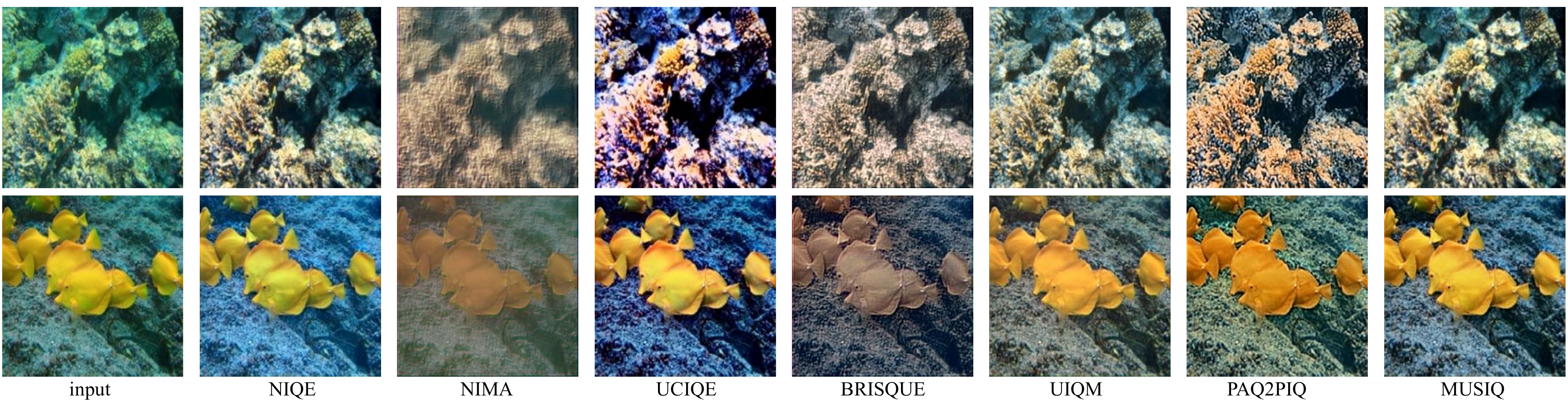}
    \vspace{-0.4cm}
    \caption{Examples of pseudo labels selected by NR-IQA metrics during training.}
    % \vspace{-0.8cm}
    \vspace{-0.4cm}
    \label{fig:real_breakdown}
\end{figure}

% \section{Influence of NR-IQA Metrics}
To compare the efficiency of NR-IQA metrics, we additionally show the performance of adopting all seven NR-IQA metrics in Table \ref{tab2}. It is easy to check that using MUSIQ can achieve the best performance. Moreover, Fig. \ref{fig:real_breakdown} presents some examples of pseudo labels (images in the reliable bank) selected by seven NR-IQA methods in the training process. It can be observed that MUSIQ can help select more visually pleasing pseudo labels (rightmost) over other metrics.
% we discussed in Sec 4.6 to investigate the influence of using three representative metrics for building reliable bank. 
\vspace{-0.1cm}
 \begin{table*}[!t]
\centering
\vspace{-0.3cm}
\caption{Evaluation the influence of adopting different NR-IQA metrics on testS\cite{Anwar2019UWCNN} and testR\cite{UIEBD} in terms of PSNR and SSIM.}
\vspace{-0.3cm}
\scalebox{0.99}{
\setlength{\tabcolsep}{2.4mm}{
\begin{tabular}{|c|c|c|c|c|c|c|c|}
\toprule
 & \textbf{NIQE} & \textbf{NIMA} &  \textbf{UCIQE} & \textbf{BRISQUE} & \textbf{UIQM} &\textbf{PAQ2PIQ} &\textbf{MUSIQ} \\ 
\hline
Reliability &13.45\% &41.05\% &48.16\% &48.69\% &76.87\% &82.11\% &91.21\%  \\
testS  &22.83/0.811 &23.01/0.815 &22.90/0.813 &23.15/0.820 &23.24/0.820 &23.08/0.818 &23.40/0.821  \\
testR &22.98/0.887 &23.88/0.888 &23.64/0.890 &24.00/0.900 &23.80/0.897 &24.28/0.893 &24.59/0.901 \\
\bottomrule
\end{tabular}}}
\label{tab2}
\end{table*} 

\subsection{Inherent Advantage of Semi-UIR}
In order to verify the inherent advantage of our proposed semi-supervised restoration framework Semi-UIR, we replace the AIM-Net with 5-layer Unet used by FUnIE-GAN\cite{EUVP-FUnieGan}, and compare the performance using and not using Semi-UIR. Training details and datasets are unchanged. The quantitative results are shown in Table \ref{tab1}. It's obvious that our semi-supervised framework Semi-UIR is beneficial to improve the generalizability of general model like Unet on real-world underwater benchmarks. It also demonstrates that Semi-UIR is extensible.
% \vspace{-0.8cm}
\begin{table*}[!t]
\centering
\small
\vspace{-0.1cm}
\caption{Evaluations on non-reference benchmarks UIEB \cite{UIEBD}, EUVP \cite{EUVP-FUnieGan}, RUIE \cite{liu2020real} and Seathru \cite{seathru} in terms of UIQM \cite{UIQM}, UCIQE \cite{UCIQE} and MUSIQ \cite{ke2021musiq}. Unet-base refers to training Unet without semi-supervised learning and unlabeled data. Unet-semi denotes training Unet with our proposed Semi-UIR.}
\vspace{-0.1cm}
\scalebox{0.95}{
\setlength{\tabcolsep}{2.4mm}{
\begin{tabular}{@{}|c|c|c|c|c|c|c|c|c|c|c|c|c|@{}}
\toprule
\multicolumn{1}{|c|}{\multirow{2}{*}{\textbf{Method}}} & \multicolumn{4}{|c|}{\cellcolor[HTML]{FFCCC9}\textbf{UIQM} (higher, better)} &  \multicolumn{4}{|c|}{\cellcolor[HTML]{9AFF99}\textbf{UCIQE} (higher, better)}&\multicolumn{4}{|c|}{\cellcolor[HTML]{5DADE2}\textbf{MUSIQ} (higher, better)}\\ 
% \midrule
& \textbf{UIEB} & \textbf{EUVP} & \textbf{RUIE} &\textbf{Seathru} &\textbf{UIEB} & \textbf{EUVP} &\textbf{RUIE} & \textbf{Seathru}&\textbf{UIEB} & \textbf{EUVP} &\textbf{RUIE} & \textbf{Seathru}\\
\midrule
Unet-base &4.215 &4.442 &4.529 &4.970 &0.585 &0.583 &0.554 &0.603 &39.33 &43.87 &31.17 &62.92\\
Unet-semi &4.329 &4.512 &4.763 &5.037 &0.586 &0.597 &0.570 &0.620 &41.06 &49.11 &34.41 &64.13\\ 
\bottomrule
\end{tabular}}}
\label{tab1}
\end{table*}
\subsection{Additional Experimental Results on Non-reference Benchmark}
\begin{figure*}[h]
\centering
\includegraphics[width=1.0\linewidth]{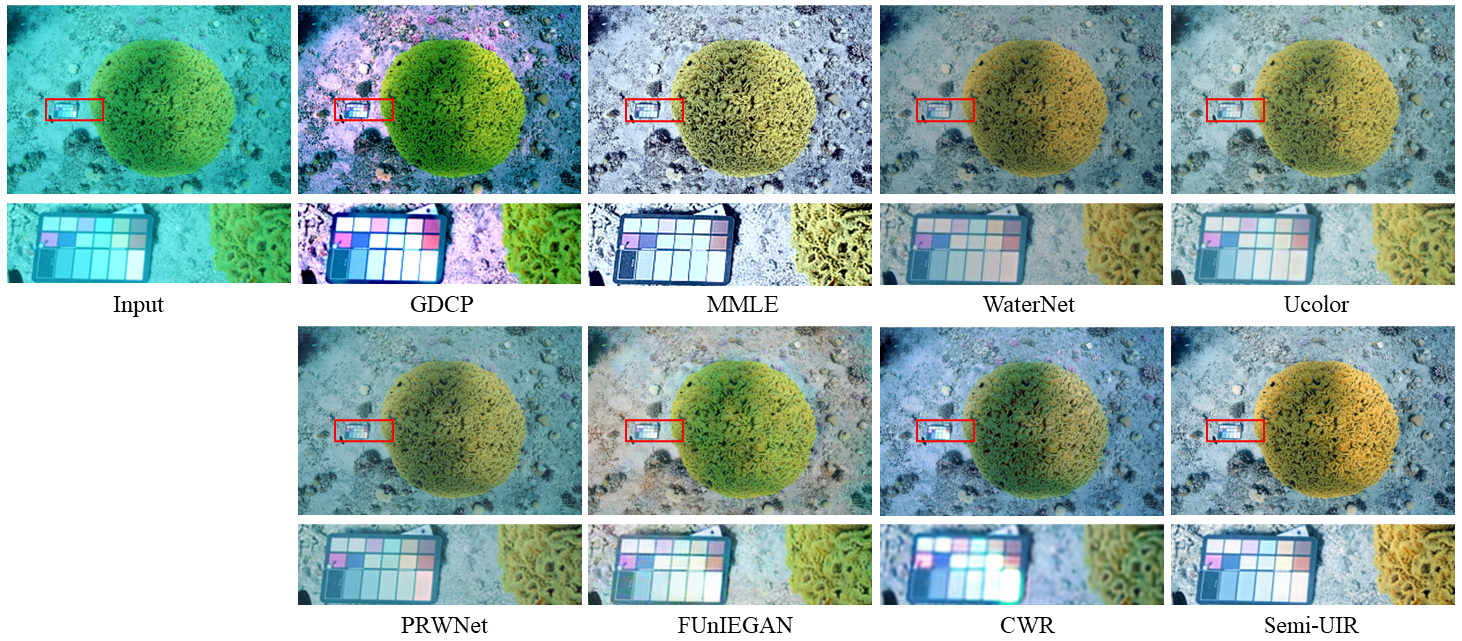}
\vspace{-0.5cm}
\caption{Visual comparisons on non-reference benchmark Seathru \cite{seathru}.}
\vspace{-0.5cm}
\label{example1}
\end{figure*}

\begin{figure*}[b]
\centering
\includegraphics[width=1.0\linewidth]{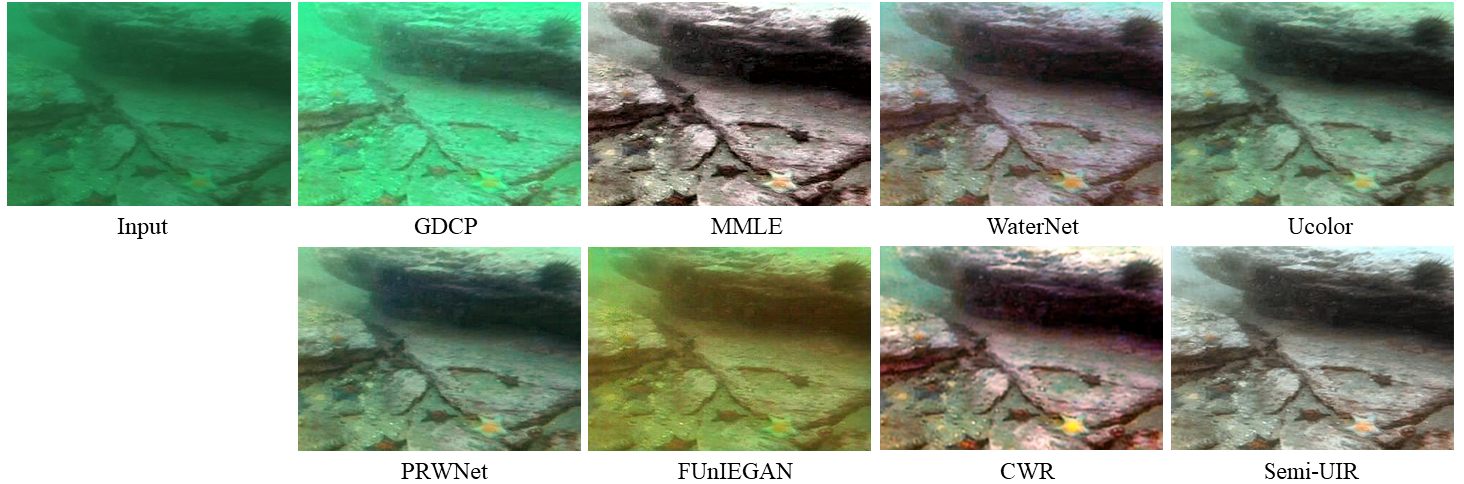}
\vspace{-0.5cm}
\caption{Visual comparisons on non-reference benchmark RUIE \cite{liu2020real}.}
\label{example2}
\end{figure*}

\begin{figure*}[b]
\centering
\includegraphics[width=1.0\linewidth]{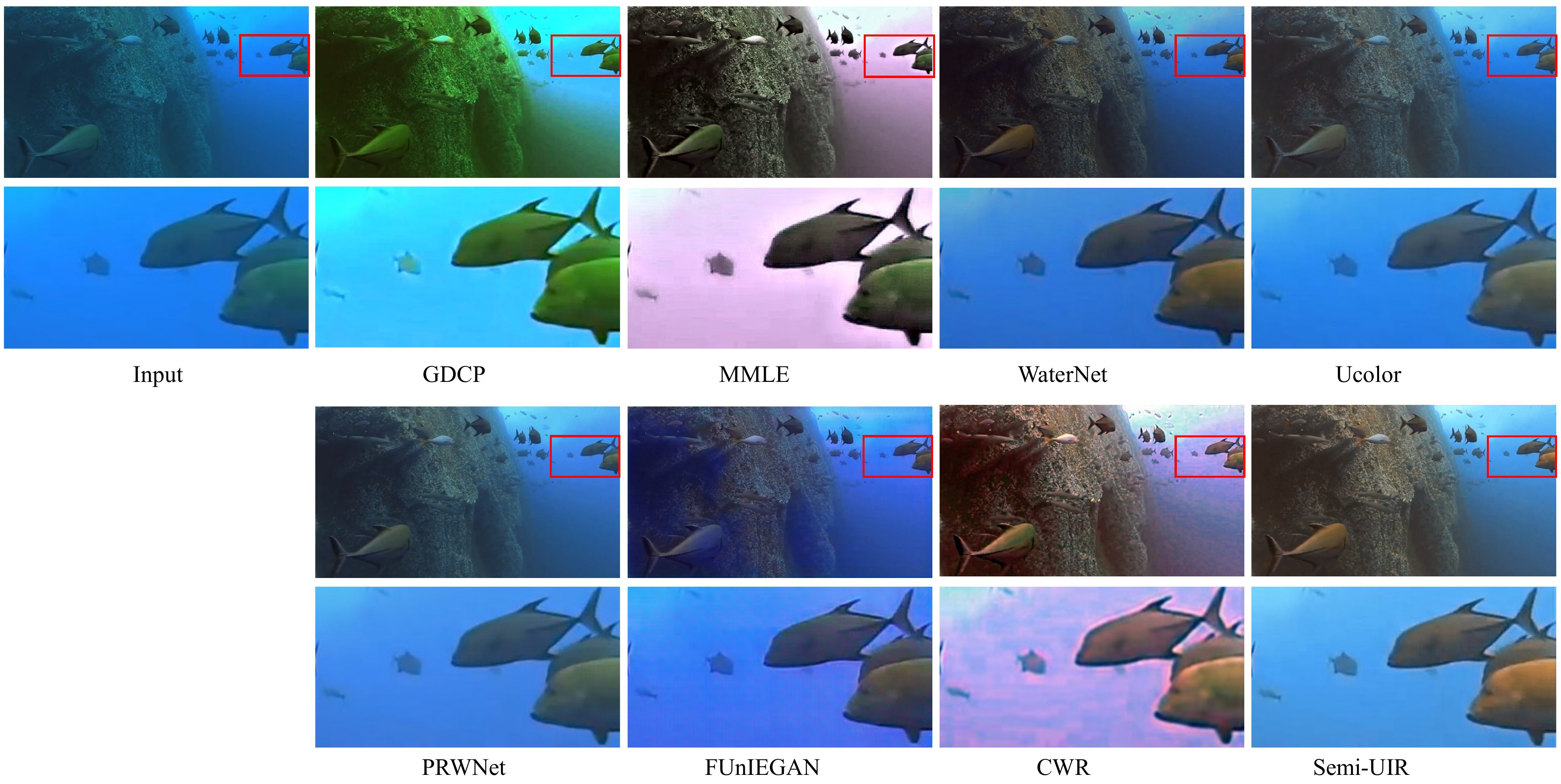}
\includegraphics[width=1.0\linewidth]{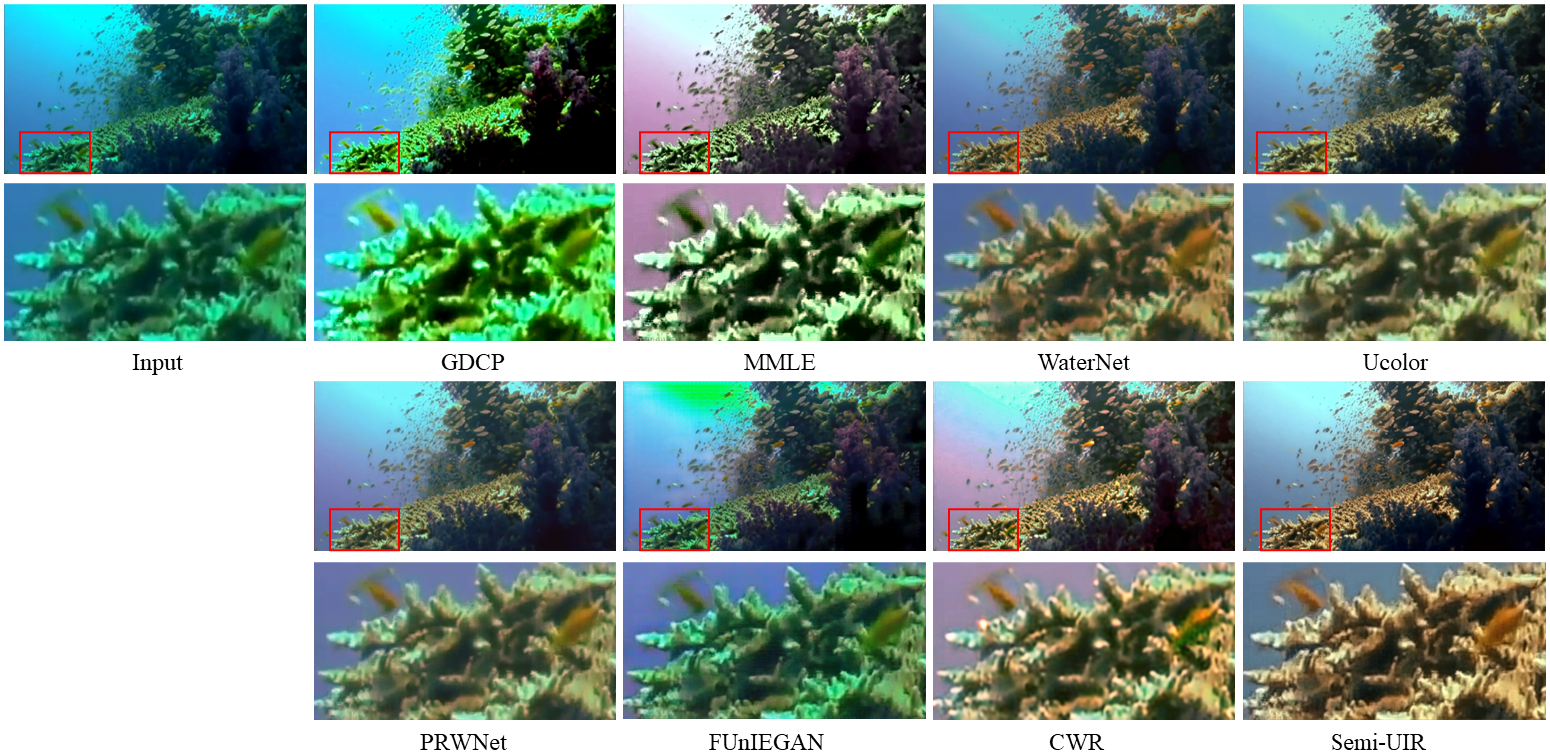}
% \vspace{-0.5cm}
\caption{Visual comparisons on non-reference benchmark UIEB \cite{UIEBD}.}
\label{example4}
\end{figure*}

In Fig. \ref{example1}- \ref{example4}, we present more results of our Semi-UIR on non-reference benchmarks Seathru \cite{seathru}, RUIE \cite{liu2020real}, UIEB \cite{UIEBD} and EUVP \cite{EUVP-FUnieGan}, and compare with the state-of-the-art methods including GDCP \cite{peng2018gdcp}, MMLE \cite{MMLE}, WaterNet \cite{UIEBD}, Ucolor \cite{ucolor}, FUnIE-GAN \cite{EUVP-FUnieGan}, PRWNet \cite{prwnet} and CWR \cite{cwr}. Our proposed Semi-UIR outperforms other algorithms in restoring underwater images with rich details and natural color. 
\vspace{-0.2cm}
\begin{figure*}[b]
\centering
\includegraphics[width=0.95\linewidth]{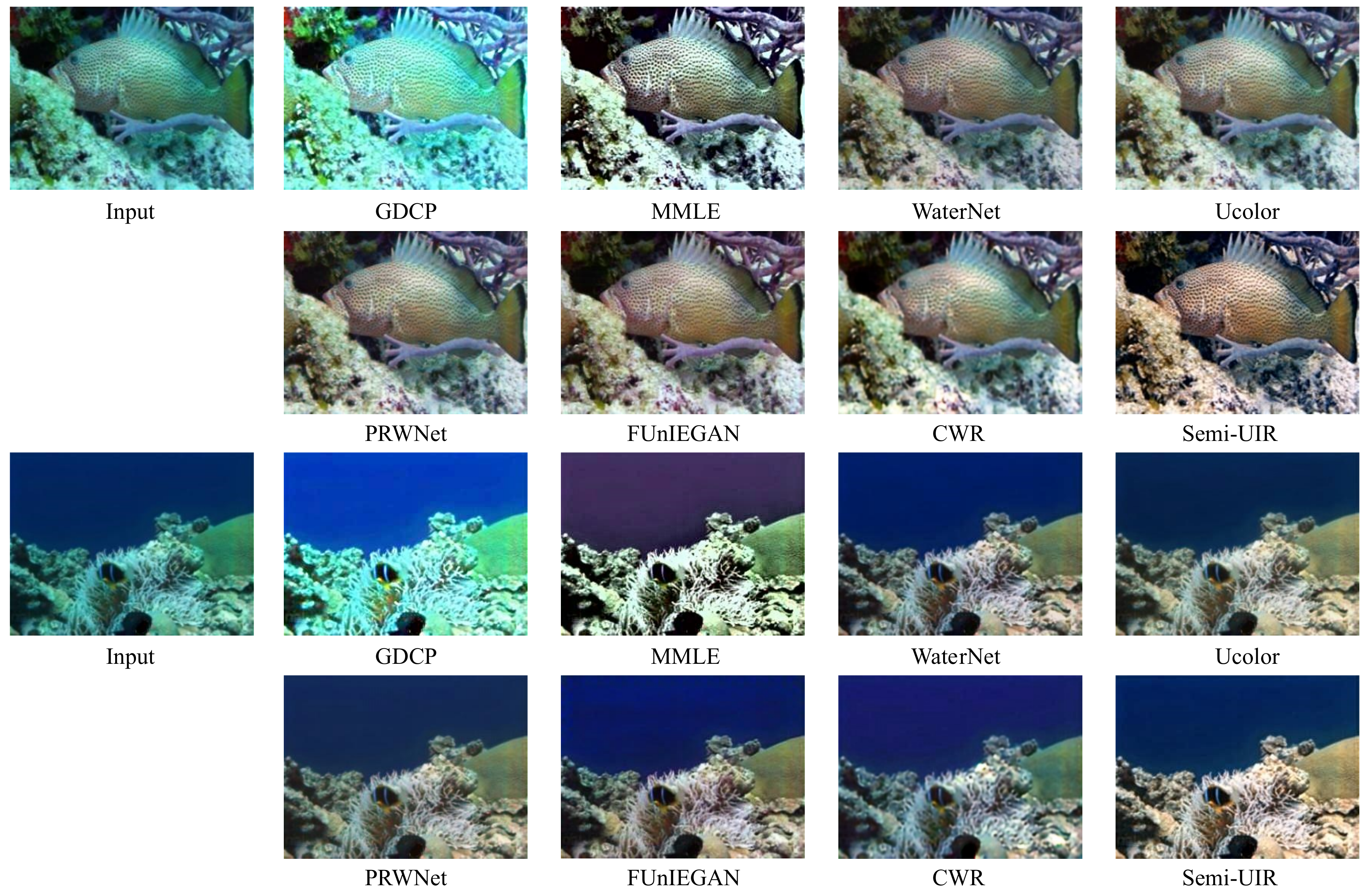}
% \vspace{-0.2cm}
\includegraphics[width=0.95\linewidth]{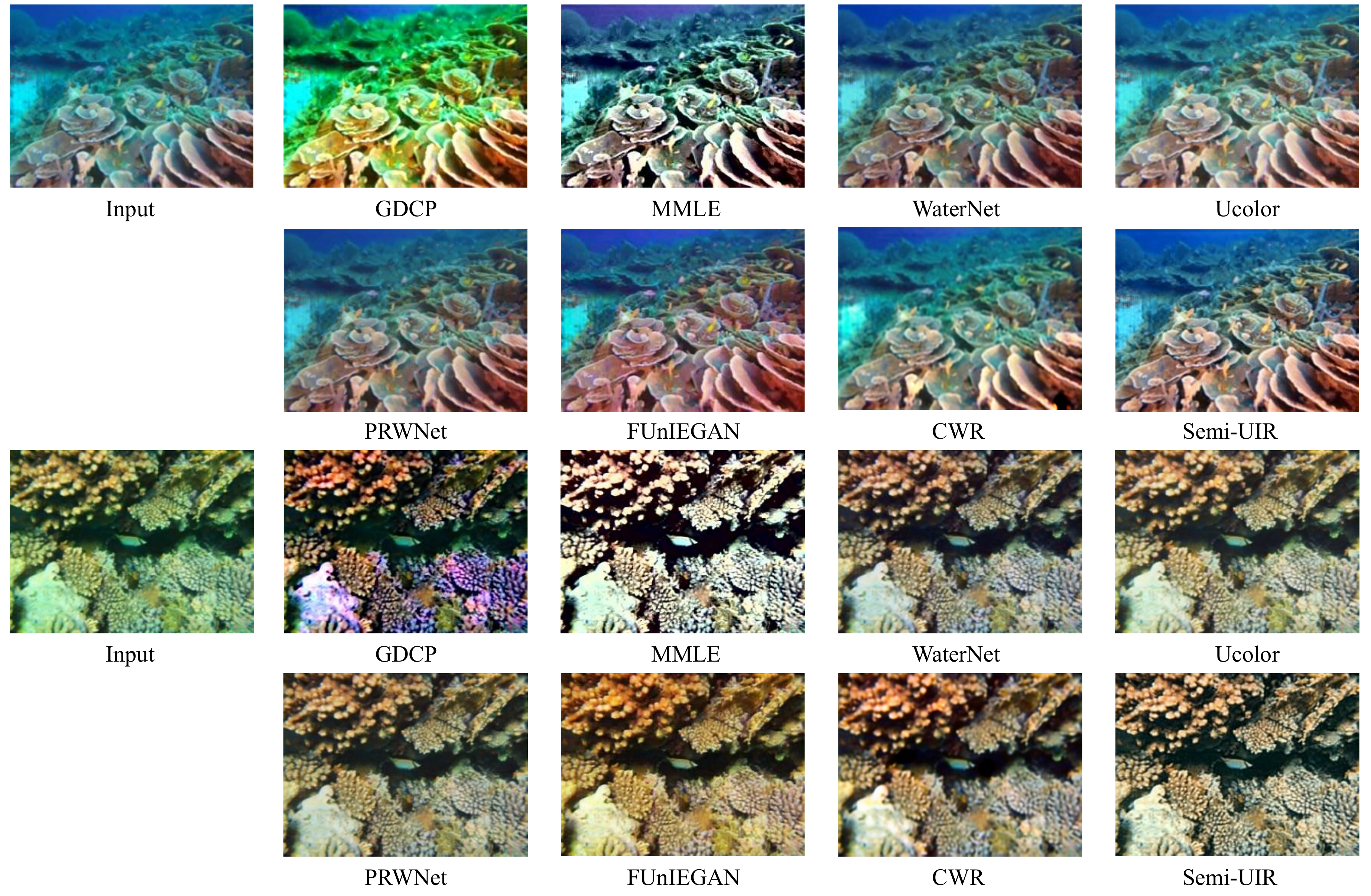}
% \vspace{-0.8cm}
\caption{Visual comparisons on non-reference benchmark EUVP \cite{EUVP-FUnieGan}.}
\label{example3}
\end{figure*} 
\vspace{0.1cm}

\end{document}